\documentclass[10pt,twocolumn,letterpaper]{article}

\usepackage[pagenumbers]{cvpr} 

\usepackage{bm}
\usepackage{tabularx}
\usepackage{algorithm,algpseudocode}
\usepackage{makecell}
\usepackage{array}
\usepackage{tikz}
\usetikzlibrary{tikzmark}
\usepackage{afterpage}
\usepackage{amssymb}
\usepackage{pifont}
\newcommand{\tabitem}{~~\llap{\textbullet}~~}
\usepackage{extarrows} 
\usepackage{nicefrac} 
\PassOptionsToPackage{dvipsnames}{xcolor}

\newcolumntype{L}{>{\raggedright\arraybackslash}X}
\usepackage{graphbox}

\definecolor{cvprblue}{rgb}{0.21,0.49,0.74}
\usepackage[pagebackref,breaklinks,colorlinks,citecolor=cvprblue]{hyperref}

\def\paperID{XXXX} 
\def\confName{XXXX}
\def\confYear{XXXX}

\title{Generative Powers of Ten}
\author{
Xiaojuan Wang\textsuperscript{1}\space\space 
Janne Kontkanen\textsuperscript{2}\space\space 
Brian Curless\textsuperscript{1, 2}\space\space 
Steven M. Seitz\textsuperscript{1, 2}\space\space
Ira Kemelmacher-Shlizerman \textsuperscript{1, 2}\space\space \\
Ben Mildenhall\textsuperscript{2}\space \space
Pratul Srinivasan\textsuperscript{2}\space \space 
Dor Verbin\textsuperscript{2}\space \space 
Aleksander Holynski\textsuperscript{2, 3}
\vspace{1mm} \\
\textsuperscript{1}University of Washington\qquad 
\textsuperscript{2}Google Research\qquad 
\textsuperscript{3}UC Berkeley \vspace{1mm}\\
\href{https://powers-of-10.github.io/}{ powers-of-ten.github.io}
}

\begin{document}

\twocolumn[{%
\renewcommand\twocolumn[1][]{#1}%
\maketitle
\begin{center}
    \centering
    \captionsetup{type=figure}
    \includegraphics[width=\textwidth]{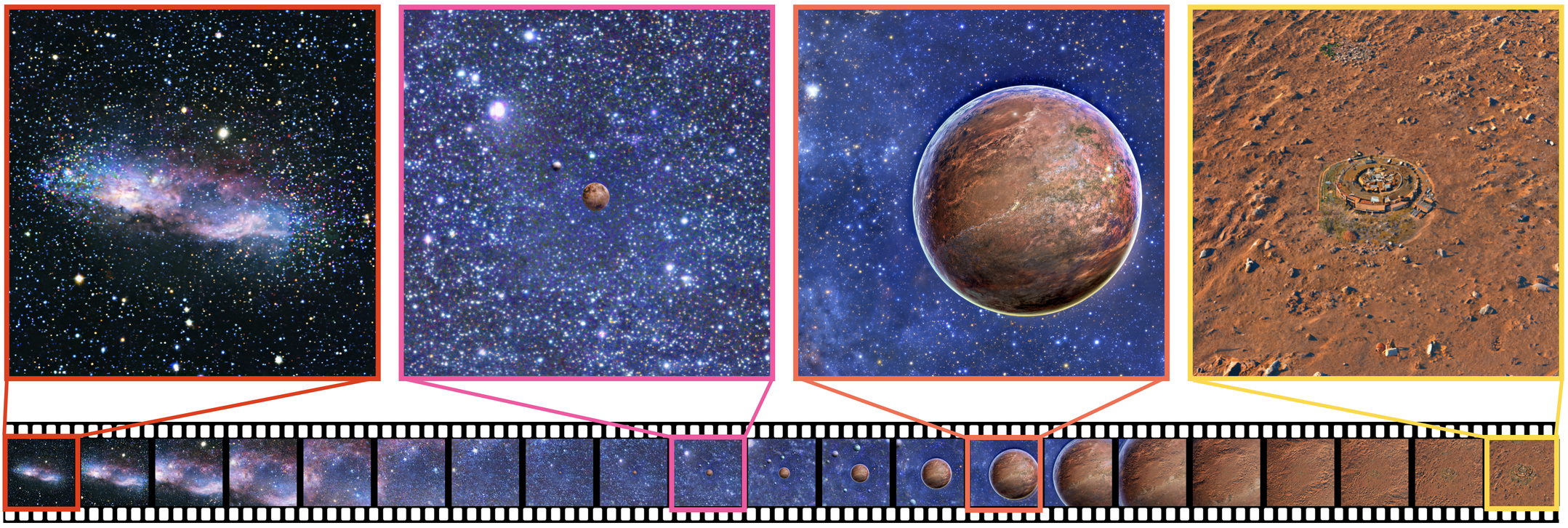}
    \captionof{figure}{Given a series of prompts describing a scene at varying zoom levels, \eg, from a distant galaxy to the surface of an alien planet, our method uses a pre-trained text-to-image diffusion model to generate a continuously zooming video sequence.}
\end{center}%
}]

\begin{abstract}
We present a method that uses a text-to-image model to generate consistent content across multiple image scales, enabling extreme {\em semantic zooms} into a scene, 
\eg ranging from a wide-angle landscape view of a forest to a macro shot of an insect sitting on one of the tree branches.
We achieve this through a joint multi-scale diffusion sampling approach that encourages consistency across different scales while preserving the integrity of each individual sampling process. Since each generated scale is guided by a different text prompt, our method enables deeper levels of zoom than traditional super-resolution methods that may struggle to create new contextual structure at vastly different scales. We compare our method 
qualitatively with alternative techniques in image super-resolution and outpainting, and show that our method is most effective at generating consistent multi-scale content. 
\end{abstract}

\vspace{-10pt}
 \section{Introduction}
\begin{figure}[t!]
    \centering
    \includegraphics[width=1.0\linewidth]{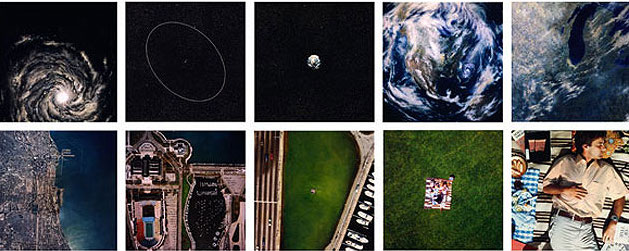}
    \caption{{\bf Powers of Ten (\textit{1977})} This documentary film illustrates the relative scale of the universe as a single shot that gradually zooms out from a human to the universe, and then back again to the microscopic molecular level.}
    \label{fig:po10}
    \vspace{-20pt}
\end{figure}

Recent advances in text-to-image models~\cite{zhang2023adding, brooks2023instructpix2pix, ruiz2023dreambooth, ruiz2023hyperdreambooth, hertz2022prompt, epstein2023diffusion, po2023state} have been transformative in enabling applications like image generation from a single text prompt.  
But while digital images exist at a fixed resolution, the real world can be experienced at many different levels of scale. Few things exemplify this better than the classic 1977 short film \textit{``Powers of Ten''}, shown in Figure~\ref{fig:po10}, which showcases the sheer magnitudes of scale that exist in the universe by visualizing a continuous zoom from the outermost depths of the galaxy to the cells inside our bodies\footnote{\url{https://www.youtube.com/watch?v=0fKBhvDjuy0}}. Unfortunately, producing animations or interactive experiences like these has traditionally required trained artists and many hours of tedious labor---and although we might want to replace this process with a generative model, existing methods have not yet demonstrated the ability to generate consistent content across multiple zoom levels.  

Unlike traditional super-resolution methods, which generate higher-resolution content conditioned on the pixels of the original image, extreme zooms expose entirely new structures, \eg, magnifying a hand to reveal its underlying skin cells.  Generating such a zoom requires {\em semantic} knowledge of human anatomy.
In this paper, we focus on solving this {\em semantic zoom} problem,
\ie, enabling text-conditioned multi-scale image generation, to create \emph{Powers of Ten}-like zoom videos. As input, our method expects a series of text prompts that describe different scales of the scene, and produces as output a multi-scale image representation that can be explored interactively or rendered to a seamless zooming video. These text prompts can be user-defined (allowing for creative control over the content at different zoom levels) or crafted with the help of a large language model (\eg, by querying the model with an image caption and a prompt like \emph{``describe what might you see if you zoomed in by 2x''}).

At its core, our method relies on a joint sampling algorithm that uses a set of parallel diffusion sampling processes distributed across zoom levels. These sampling processes are coordinated to be consistent through an iterative frequency-band consolidation process, in which intermediate image predictions are consistently combined across scales.  Unlike existing approaches that accomplish similar goals by repeatedly increasing the effective image resolution (\eg, through super-resolution or image outpainting), our sampling process jointly optimizes for the content of all scales at once, allowing for both (1) plausible images at each scale and (2) consistent content across scales. Furthermore, existing methods are limited in their ability to explore wide ranges of scale, since they 
rely primarily on the input image content to determine the added details at subsequent zoom levels. In many cases, image patches contain insufficient contextual information to inform detail at deeper (\eg, 10x or 100x) zoom levels. On the other hand, our method grounds each scale in a text prompt, allowing for new structures and content to be conceived across extreme zoom levels. In our experiments, we compare our work qualitatively to these existing methods, and demonstrate that the zoom videos that our method produces are notably more consistent. Finally, we showcase a number of ways in which our algorithm can be used, \eg, by conditioning purely on text or grounding the generation in a known (real) image. 

\section{Prior Work}
{\noindent \bf Super-resolution and inpainting.} Existing text-to-image based super resolution models~\cite{saharia2022image, stabilityai} and outpainting models~\cite{stabilityai, saharia2022palette, ramesh2022hierarchical, tang2023realfill} can be adapted to the zoom task as autoregressive processes, \ie, by progressively outpainting a zoomed-in image, or progressively super-resolving a zoomed-out image. One significant drawback of these approaches is that 
later-generated images have no influence on the previously generated ones, which can often lead to suboptimal results, as certain structures may be entirely incompatible with subsequent levels of detail, causing error accumulation across recurrent network applications.

\vspace{0.1cm}
{\noindent \bf Perpetual view generation.} Starting from a single view RGB image, perpetual view generation methods like Infinite Nature~\cite{li2022infinitenature} and InfiniteNature-Zero~\cite{liu2021infinite} learn to generate unbounded flythrough videos of natural scenes. These methods differ from our generative zoom in two key ways: (1) they translate the camera in 3D, causing a ``fly-through'' effect with perspective effects, rather than the ``zoom in" our method produces, and (2) they synthesize the fly-through starting from a single image by progressivly inpainting unknown parts of novel views, wheras we generate the entire zoom sequence simultaneously and coherently across scales, with text-guided semantic control.

{\vspace{ 0.1cm}}
{\noindent \bf Diffusion joint sampling for consistent generation.} Recent research~\cite{bar2023multidiffusion, zhang2023diffcollage, tang2023mvdiffusion, lee2023syncdiffusion} leverages pretrained diffusion models to generate arbitrary-sized images or panoramas from smaller pieces using joint diffusion processes. These processes involve concurrently generating these multiple images  by merging their intermediate results within the sampling process. In particular, {\it DiffCollage}~\cite{zhang2023diffcollage} introduces a factor graph formulation to express spatial constraints among these images, representing each image as a node, and overlapping areas with additional nodes. Each sampling step involves aggregating individual predictions 
based on the factor graph.
For this to be possible, a given  diffusion model needs to be finetuned for different factor nodes. Other works such as {\it MultiDiffusion}~\cite{bar2023multidiffusion} reconciles different denoising steps by solving for a least squares optimal solution: \ie, averaging the diffusion model predictions at overlapping areas. However, none of these approaches can be applied to our problem, where our jointly sampled images have spatial correspondence at vastly different spatial scales.

\algblock{ParFor}{EndParFor}
\algnewcommand\algorithmicparfor{\textbf{parfor}}
\algnewcommand\algorithmicpardo{\textbf{do}}
\algnewcommand\algorithmicendparfor{\textbf{end\ parfor}}
\algrenewtext{ParFor}[1]{\algorithmicparfor\ #1\ \algorithmicpardo}
\algrenewtext{EndParFor}{\algorithmicendparfor}

\section{Preliminaries}
Diffusion models~\cite{sohl2015deep,song2019generative, song2020denoising, song2020score,ho2020denoising,dhariwal2021diffusion} generate images from random noise through a sequential sampling process. This sampling process reverses a destructive process that gradually adds Gaussian noise on a clean image $\mathbf{x}$. The intermediate noisy image at time step $t$ is expressed as:
\begin{equation*}
    \mathbf{z}_t = \alpha_t\mathbf{x} + \sigma_t\bm{\epsilon}_t,
\end{equation*}
where $\bm{\epsilon}_t\sim \mathcal{N}(\mathbf{0, I})$ is a standard Gaussian noise, and $\alpha_t$ and $\sigma_t$ define a fixed noise schedule, with larger $t$ corresponding to more noise. A diffusion model is a neural network $\bm{\epsilon}_{\theta}$ that predicts the approximate clean image $\hat{\bf{x}}$ directly, or equivalently the added noise $\bm{\epsilon}_t$ in $\mathbf{z}_t$. The network is trained with the loss:
\begin{equation*}
 \mathcal{L}(\theta) = \mathbb{E}_{t\sim U[1, T], \bm{\epsilon}_t\sim \mathcal{N}(\mathbf{0}, \mathbf{I})}[w(t)\|\bm{\epsilon}_{\theta}(\mathbf{z}_t; t, y)-\bm{\epsilon}_t\|_2^2],
\end{equation*}
where $y$ is an additional conditioning signal like text~\cite{saharia2022photorealistic, ramesh2022hierarchical, rombach2022high}, and $w(t)$ is a weighting function typically set to 1~\cite{ho2020denoising}. A standard choice for $\bm{\epsilon}_{\theta}$ is a
U-Net with self-attention and cross-attention operations attending to the conditioning $y$.

Once the diffusion model is trained, various sampling methods~\cite{ho2020denoising, song2020denoising, liu2022pseudo} are designed to sample efficiently from the model, starting from pure noise $\mathbf{z}_T\sim \mathcal{N}(\mathbf{0}, \mathbf{I})$ and iteratively denoising it to a clean image. These sampling methods often rely on classifier-free guidance~\cite{ho2020denoising}, a process which uses a linear combination of the text-conditional and unconditional predictions to achieve better adherence to the conditioning signal:
\begin{equation*}
\hat{\bm{\epsilon}}_t =(1+\omega)\bm{\epsilon}_{\theta}(\mathbf{z}_t; t, y) - \omega\bm{\epsilon}_{\theta}(\mathbf{z}_t; t).
\end{equation*}
This revised $\hat{\bm{\epsilon}}_t$ is used as the noise prediction to update the noisy image $\mathbf{z}_t$. Given a noisy image and a noise prediction, the estimated clean image $\hat{\mathbf{x}}_t$ is computed as $ \hat{\mathbf{x}}_t = (\mathbf{z}_t-\sigma_t\hat{\bm{\epsilon}}_t)/\alpha_t$.
The iterative update function in the sampling process depends on the sampler used; in this paper we use DDPM~\cite{ho2020denoising}.

\section{Method}
Let $y_0, ..., y_{N-1}$ be a series of prompts describing a single scene at varying, corresponding zoom levels $p_0, ..., p_{N-1}$ forming a geometric progression, \ie, $p_i = p^i$
(we typically set $p$ to $2$ or $4$). Our objective is to generate a sequence of corresponding $H\times W\times C$ images
$\mathbf{x}_0, ..., \mathbf{x}_{N-1}$ from an existing, pre-trained, text-to-image diffusion model. We aim to generate the entire set of images jointly in a zoom-consistent way. This means that the image $\mathbf{x}_{i}$ at any specific zoom level $p_i$, should be consistent with the center $H/p\times W/p$ crop of the zoomed-out image $\mathbf{x}_{i-1}$.

We propose a {\it multi-scale joint sampling} approach and a corresponding {\it zoom stack} representation that gets updated in the diffusion-based sampling process. In Sec.~\ref{sec:representation}, we introduce our zoom stack representation and the process that allows us to render it into an image at any given zoom level. In Sec.~\ref{sec:lap blending}, we present an approach for consolidating multiple diffusion estimates into this representation in a consistent way. Finally, in Sec.~\ref{sec:multi_scale_Sample}, we show how these components are used in the complete sampling process.

\begin{figure}
    \centering
    \includegraphics[width=\linewidth]{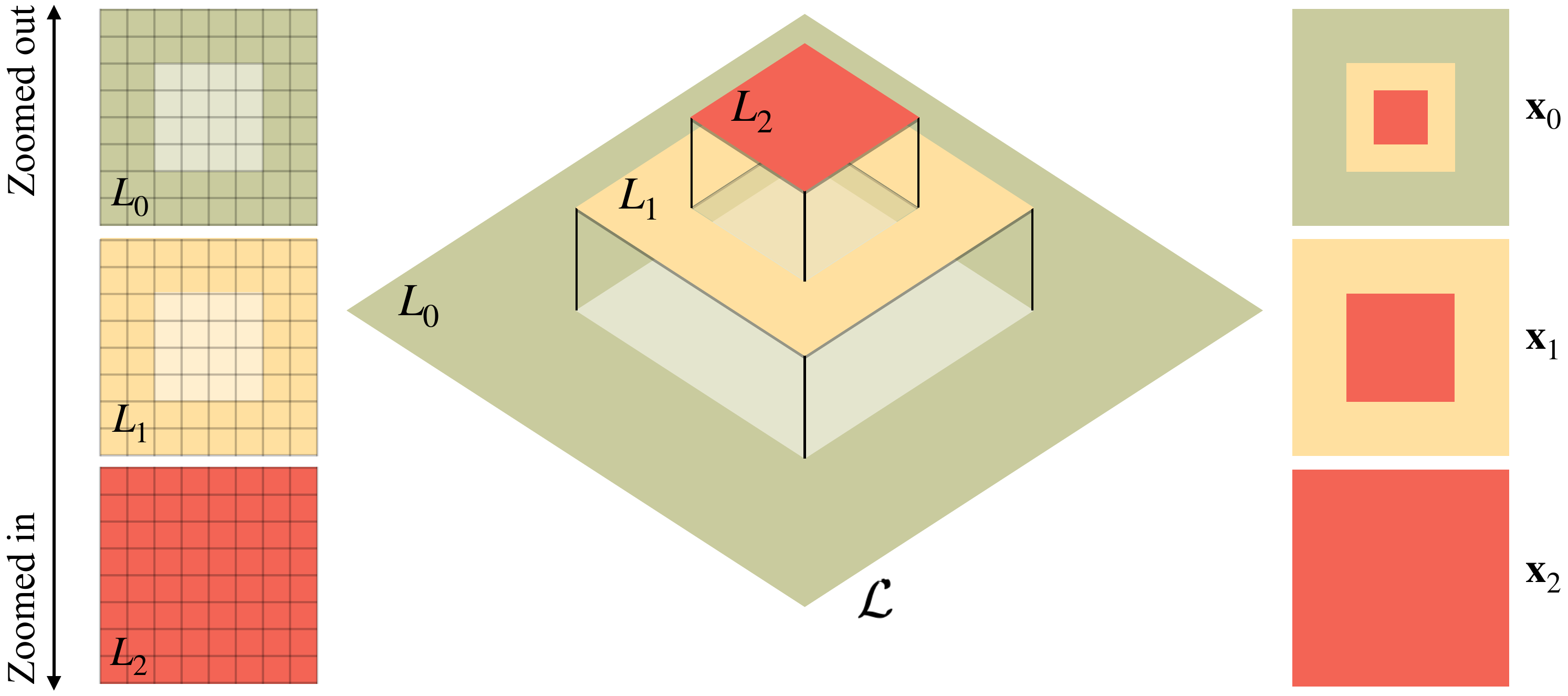}
    \caption{\textbf{Zoom stack.} Our representation consists of $N$ layer images $L_i$ of constant resolution (left). These layers are arranged in a pyramid-like structure, with layers representing finer details corresponding to a smaller spatial extent (middle). These layers are composited to form an image at any zoom level (right).}
    \label{fig:representation}
    \vspace{-10pt}
\end{figure}

\begin{figure*}[ht!]
    \centering
    \includegraphics[width=1.0\linewidth]{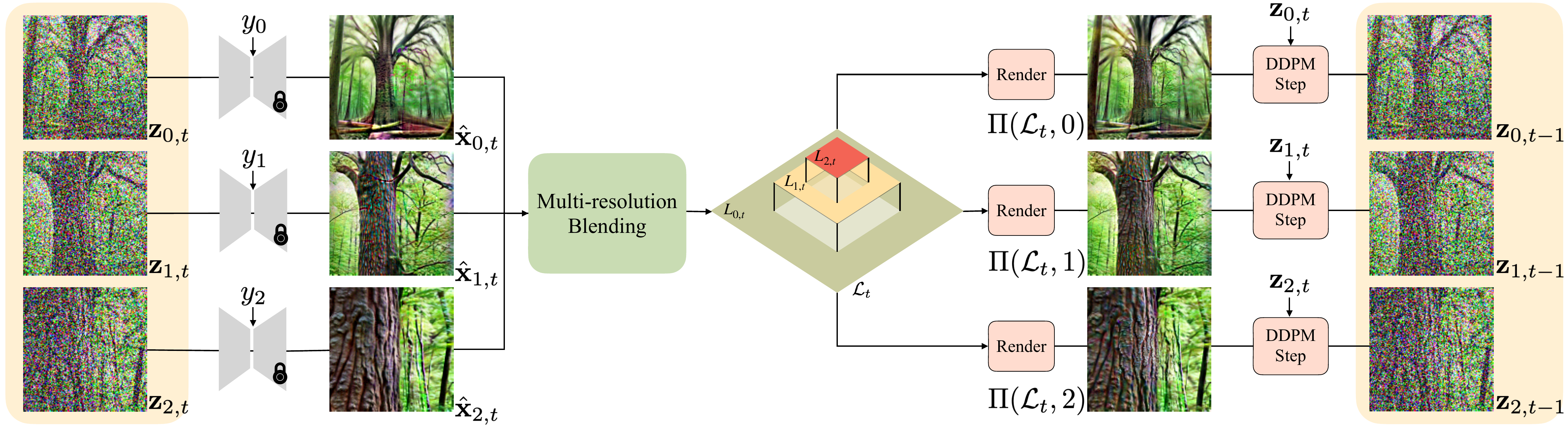}
    \caption{{\bf Overview of a single sampling step.} (1) Noisy images $\mathbf{z}_{i, t}$ from each zoom level, along with the respective prompts $y_i$ are simultaneously fed into the same pretrained diffusion model, returning estimates of the corresponding clean images $\hat{\mathbf{x}}_{i, t}$. These images may have inconsistent estimates for the overlapping regions that they all observe. We employ {\it multi-resolution blending} to fuse these regions into a consistent zoom stack $\mathcal{L}_{t}$ and re-render the different zoom levels from the consistent representation. These re-rendered images $\Pi_{\text{image}}(\mathcal{L}_t; i)$ are then used as the clean image estimates in the DDPM sampling step.}
    \label{fig:demo}
    \vspace{-10pt}
\end{figure*}

\subsection{Zoom Stack Representation}\label{sec:representation}
Our zoom stack representation, which we denote by $\mathcal{L} = (L_0, ..., L_{N-1})$, is designed to allow rendering images at any zoom level $p_0,...,p_{N-1}$.
The representation, illustrated in Fig.~\ref{fig:representation}, contains 
$N$ images of shape $H\times W$, one for each zoom level, where the $i$th image $L_i$ stores the pixels corresponding to the $i$th zoom level $p_i$.

\vspace{0.1cm}
{\noindent\bf  Image rendering.} The rendering operator, which we denote by $\Pi_{\text{image}}(\mathcal{L}; i)$, takes a zoom stack $\mathcal{L}$ and returns the image at the $i$th zoom level $p_i = p^i$. We denote by $\mathcal{D}_{i}(\mathbf{x})$ the operator which downscales the image $\mathbf{x}$ by factor $p_i$, and zero-pads the image back to size $H\times W$; and we denote by $M_i$ the corresponding $H\times W$ binary image which has value $1$ at the center $H/p_{i}\times W/p_i$ patch and value $0$ at padded pixels.
The operator $\mathcal{D}_{i}$ operates by prefiltering the image with a truncated Gaussian kernel of size $p_i\times p_i$  and resampling with a stride of $p_i$. As described in Alg.~\ref{alg:rendering}, an image $\mathbf{x}_i$ at the $i$th zoom level is rendered by starting with $L_i$, and iteratively replacing its central $H/p_j \times W/p_j$ crop with $\mathcal{D}_{j-i}(L_{j})$, for $j=i+1,...,N-1$. (In Alg.~\ref{alg:rendering} we denote by $\odot$ the elementwise multiplication of a binary mask $M$ with an image.) This process guarantees that rendering at different zoom levels will be consistent at overlapping central regions.

\vspace{0.1cm}
{\noindent\bf Noise rendering.} At every denoising iteration of DDPM~\cite{ho2020denoising}, each pixel is corrupted by globally-scaled i.i.d.\ Gaussian noise $\bm{\epsilon}\sim \mathcal{N}(\mathbf{0, I})$. Since we would like images rendered at different zoom levels to be consistent, 
it is essential to make sure the added noise is also consistent, with 
overlapping region across different zoom levels sharing the same noise structure. Therefore, we use a rendering operator similar to $\Pi_{\text{image}}$ which converts a set of independent noise images, $\mathcal{E} = (E_0, ..., E_{N-1})$ into a single zoom-consistent noise $\bm{\epsilon}_i = \Pi_{\text{noise}}(\mathcal{E}; i)$. However, because downsampling involves prefiltering, which modifies the statistics of the resulting noise, we upscale the $j$th downscaled noise component by $\nicefrac{p_j}{p_i}$ to preserve the variance, ensuring that the noise satisfies the standard Gaussian distribution assumption, \ie, that $\bm{\epsilon}_i=\Pi_{\text{noise}}(\mathcal{E}; i)\sim \mathcal{N}(\mathbf{0, I})$ for all levels $i$. 

\begin{algorithm}[h!] 
\caption{Image and noise rendering at scale $i$.}
\label{alg:rendering}
\begin{algorithmic}[1]
\Statex
\State {Set $\mathbf{x}\leftarrow L_i, \bm{\epsilon}\sim\mathcal{N}(\mathbf{0}, \mathbf{I})$}
    \For{$j=i+1,\dots, N-1$}     
        \State{$\mathbf{x} \leftarrow M_{j-i} \odot \mathcal{D}_{j-i}(L_j) + (1 - M_{j-i})\odot \mathbf{x}$}
        \State{$\bm{\epsilon} \leftarrow (p_{j}/p_{i})  M_{j-i}\odot \mathcal{D}_{j-i}(E_j) + (1 - M_{j-i})\odot \bm{\epsilon}$}
    \EndFor
    \State \Return {$\mathbf{x}, \bm{\epsilon}$}
\end{algorithmic}
\end{algorithm}

\subsection{Multi-resolution blending}\label{sec:lap blending}
Equipped with a method for rendering a zoom stack and sampling noise at any given zoom level, we now describe a mechanism for integrating multiple observations of the same scene $\mathbf{x}_0, ..., \mathbf{x}_{N-1}$ at varying zoom levels $p_0, ..., p_{N-1}$ into a consistent zoom stack $\mathcal{L}$. This process is a necessary component of the consistent sampling process, as the diffusion model applied at various zoom levels will produce inconsistent content in the overlapping regions. Specifically, the $j$th zoom stack level $L_j$ is used in rendering multiple images at all zoom levels $i \leq j$, and therefore its value should be consistent with multiple image observations (or diffusion model samples), namely $\{\mathbf{x}_i : i \leq j\}$. The simplest possible solution to this is to naïvely average the overlapping regions across all observations. This approach, however, results in blurry zoom stack images, since coarser-scale observations of overlapping regions contain fewer pixels, and therefore only lower-frequency information.

\begin{figure}[h!]
    \centering
    \includegraphics[width=1.0\linewidth]{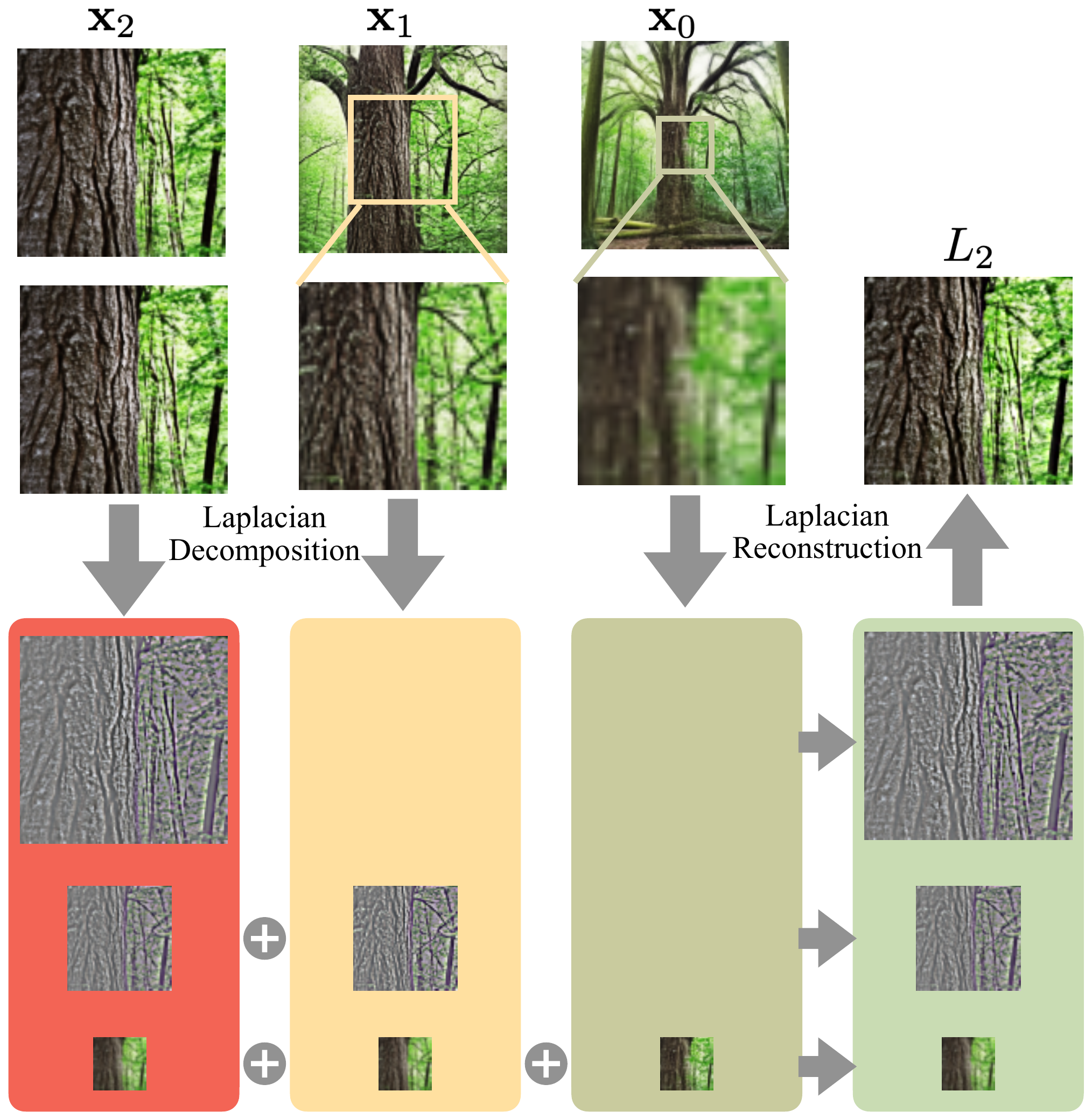}
    \caption{{\bf Multi-resolution blending.} We produce a consistent estimate for Layer $L_i$ in the zoom stack by merging the $H/p_j\times W/p_j$ central region of the corresponding zoomed out images $\mathbf{x}_{j}$ for $j \leq i$.
    This merging process involves (1) creating a Laplacian pyramid from each observation, and blending together the corresponding frequency bands to create a blended pyramid. This blended pyramid is recomposed into an image, which is used to update the layer $L_i$.}
    \label{fig:lapblending}
     \vspace{-20pt}
\end{figure}

To solve this, we propose an approach we call {\it multi-resolution blending}, which uses Laplacian pyramids to selectively fuse the appropriate frequency bands of each observation level, which prevents aliasing as well as over-blurring. 
We show an outline of this process in Fig.~\ref{fig:lapblending}.
More concretely, to update the $i$th layer in the zoom stack, we begin by cropping all samples $j \geq i$ to match with the content of the $i$th level, and rescaling them back to $H\times W$. We then analyze each of these $N-i-1$ images into a Laplacian pyramid~\cite{burt1987laplacian}, and average across corresponding frequency bands (see Figure~\ref{fig:lapblending}), resulting in an average Laplacian pyramid, which can be recomposed into an image and assigned to the $i$th level of the zoom stack. This process is applied for each layer of the zoom stack $L_i$, collecting from all further zoomed-out levels $j \geq i$. 

\subsection{Multi-scale consistent sampling}\label{sec:multi_scale_Sample}
Our complete {\it multi-scale joint sampling} process is shown in Alg.~\ref{alg:multiscale}. Fig.~\ref{fig:demo} illustrates a single sampling step $t$: Noisy images $\mathbf{z}_{i, t}$ in each zoom level along with the respective prompt $y_i$ are fed into the pretrained diffusion model in parallel to predict the noise $\hat{\bm{\epsilon}}_{i, t-1}$, and thus to compute the estimated clean images $\hat{\mathbf{x}}_{i, t}$. Equipped with our {\it multi-resolution blending} technique, the clean images are consolidated into a {\it zoom stack}, which is then rendered at all zoom levels, yielding consistent images $\Pi_{\text{image}}(\mathcal{L}_t; i)$. These images are then used in a DDPM update step along with the input  $\mathbf{z}_t$ to compute the next $\mathbf{z}_{t-1}$.

\begin{algorithm}[ht!] 
\caption{Multi-scale joint sampling.}
\label{alg:multiscale}
\begin{algorithmic}[1]
\Statex
\State {Set $\mathcal{L}_T\leftarrow \mathbf{0}$, $\mathbf{z}_{i, T} \sim \mathcal{N}(\mathbf{0}, \mathbf{I})$, $\forall i = 0, ..., N-1$}
    \For{$t=T,\dots, 1$}     
        \State{$\mathcal{E}\sim \mathcal{N}(\mathbf{0}, \mathbf{I})$}
        \ParFor{$i=0,\dots,N-1$}
            \State{$\mathbf{x}_{i,t} = \Pi_{\text{image}}(\mathcal{L}_t; i)$}
            \State{$\bm{\epsilon}_i = \Pi_{\text{noise}}(\mathcal{E}; i)$}
            \State{$\mathbf{z}_{i,t-1} =$ DDPM\_update$(\mathbf{z}_{i,t}, \mathbf{x}_{i,t},  \bm{\epsilon}_i)$}
            \State{$\hat{\bm{\epsilon}}_{i, t-1} = (1+\omega)\bm{\epsilon}_{\theta}(\mathbf{z}_{i, t-1}; t-1, y_i)$\\
            \quad \quad \quad \quad \quad \quad \quad $-\omega \bm{\epsilon}_{\theta}(\mathbf{z}_{i, t-1}; t-1)$}
            \State{$\hat{\mathbf{x}}_{i, t-1} = (\mathbf{z}_{i, t-1}-\sigma_{t-1}\hat{\bm{\epsilon}}_{i,t-1})/\alpha_{t-1}$}
        \EndParFor
        \State{$\mathcal{L}_{t-1} \leftarrow $ Blending$(\{\hat{\mathbf{x}}_{i, t-1}\}_{i=0}^{N-1})$}
    \EndFor
    \State \Return {$\mathcal{L}_0$}
\end{algorithmic}
\end{algorithm}

\begin{figure}
    \centering
    \def\imW{0.22\textwidth}
    \setlength{\tabcolsep}{1pt}
    \renewcommand\cellset{\renewcommand\arraystretch{0}}%
    \renewcommand{\arraystretch}{0.5}
     $\left.
    \begin{tabular}{cc}
        \includegraphics[width=\imW]{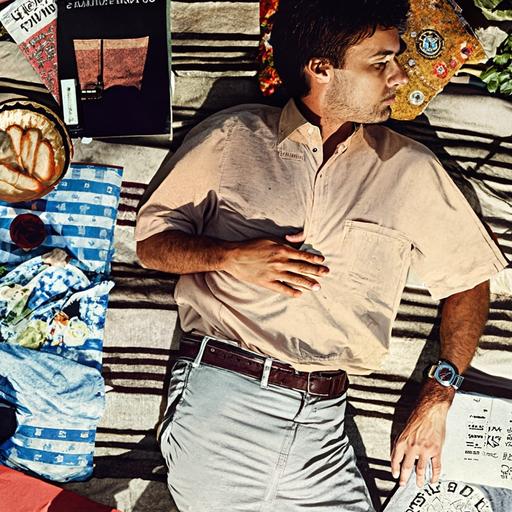} 
        &\includegraphics[width=\imW]{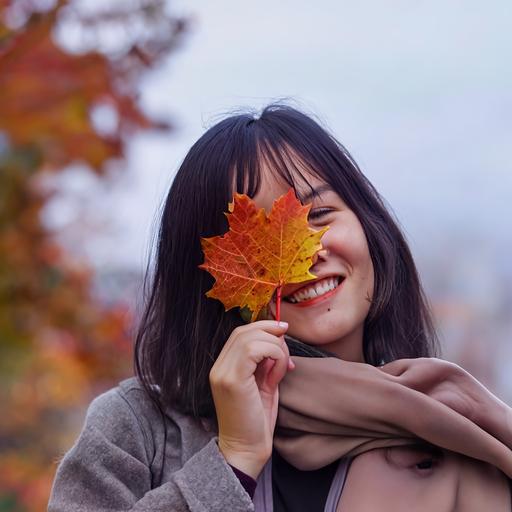} \\
        \makecell{\tiny An aerial view of a man lying on the picnic blanket}
        &\makecell{\tiny A girl is holding a maple leaf in front of her face} \\
         \includegraphics[width=\imW]{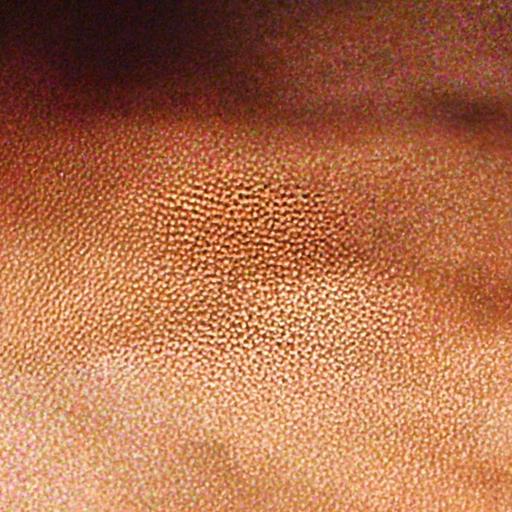} 
        &\includegraphics[width=\imW]{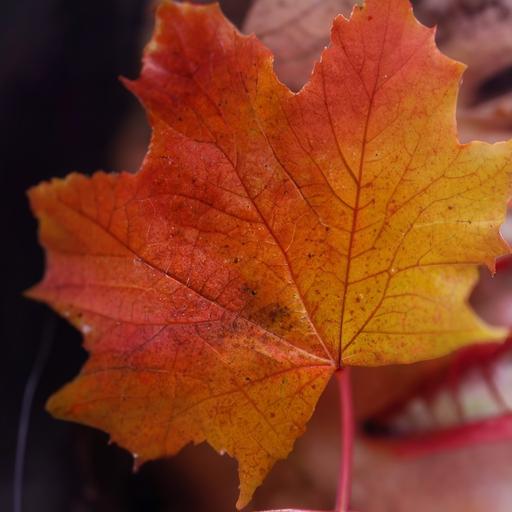} \\
        \makecell{\tiny A closeup of the surface of skin of the back hand} 
        &\makecell{\tiny A brightly colored autumn maple leaf} \\
        \includegraphics[width=\imW]{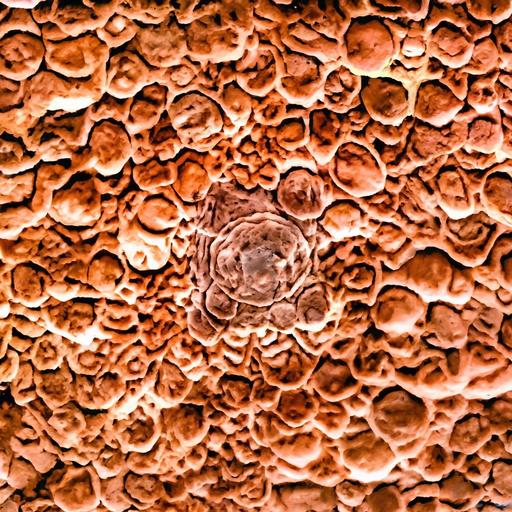} 
        &\includegraphics[width=\imW]{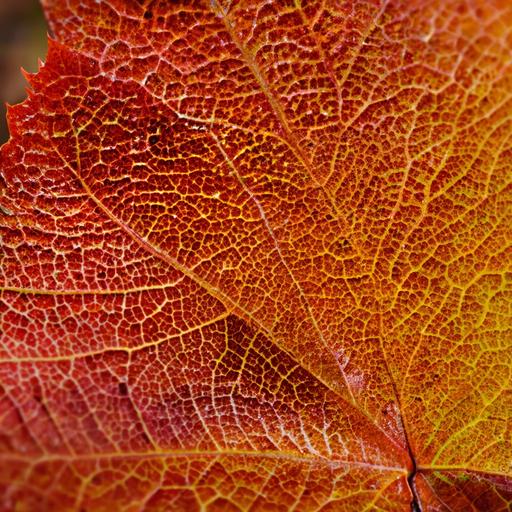} \\
        \makecell{\tiny Epidermal layer of multiple rows of tiny skin cells} 
         &\makecell{\tiny Orange maple leaf texture with lots of veins} \\
          \includegraphics[width=\imW]{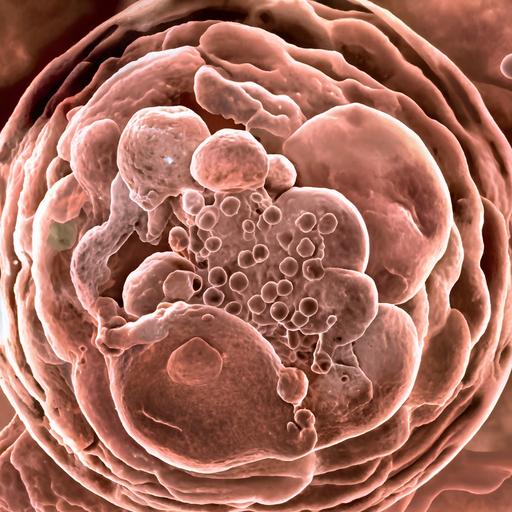} 
        &\includegraphics[width=\imW]{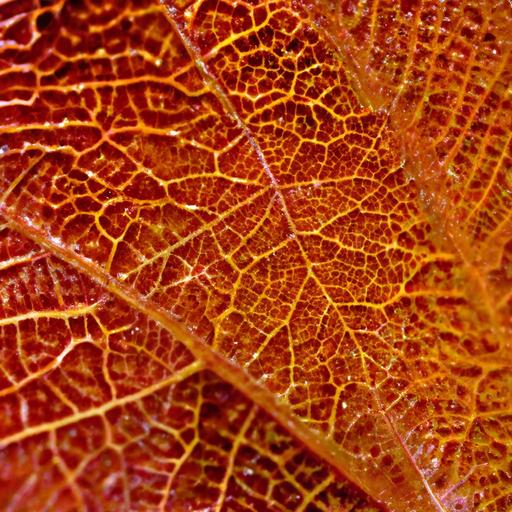} \\
        \makecell{\tiny A single round skin cell with its nucleus}
         &\makecell{\tiny Macrophoto of the veins pattern on the maple leaf}\\
          \includegraphics[width=\imW]{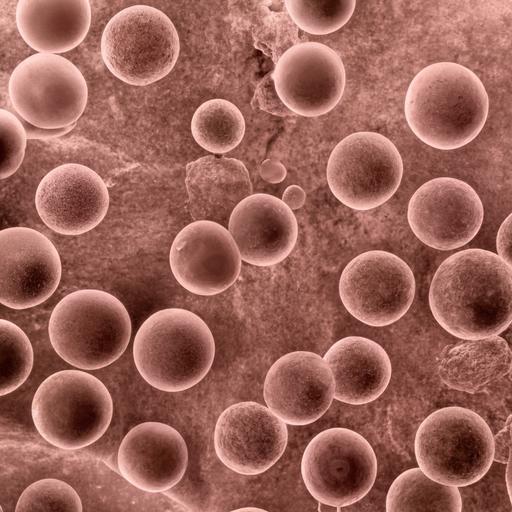} 
        &\includegraphics[width=\imW]{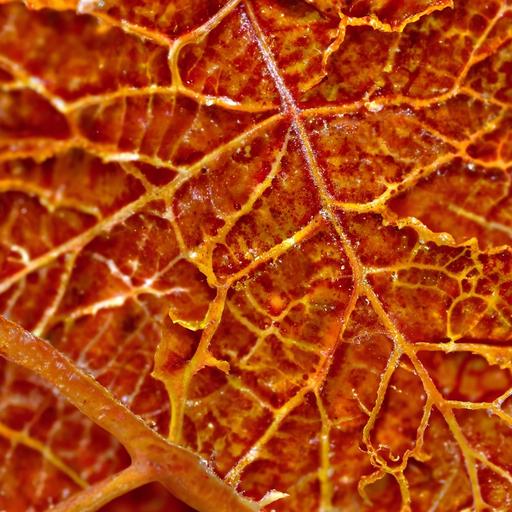} \\
        \makecell{\tiny A nucleus within a skin cell}
        &\makecell{\tiny Magnified veins pattern on the maple leaf} \\
    \end{tabular}
    \right\updownarrow
   \rotatebox[origin=c]{270}{{\scriptsize Zoomed out  \hspace{182mm} Zoomed in}}$
    \caption{ Selected images of our generated zoom sequences beginning with a provided real image. Left: Zoom from a man on a picnic blanket into the skin cells on his hand. Right: Zoom from a girl holding a leaf into the intricate vein patterns on the leaf. Face is blurred for anonymity.
    }\label{fig:real_vis}
\end{figure}

\begin{figure*}
    \centering
    \def\imW{0.23\textwidth}
    \setlength{\tabcolsep}{1pt}
    \renewcommand\cellset{\renewcommand\arraystretch{0}}%
    \renewcommand{\arraystretch}{0.5}
     $\left.
        \begin{tabular}{cccc}
            \includegraphics[width=\imW]{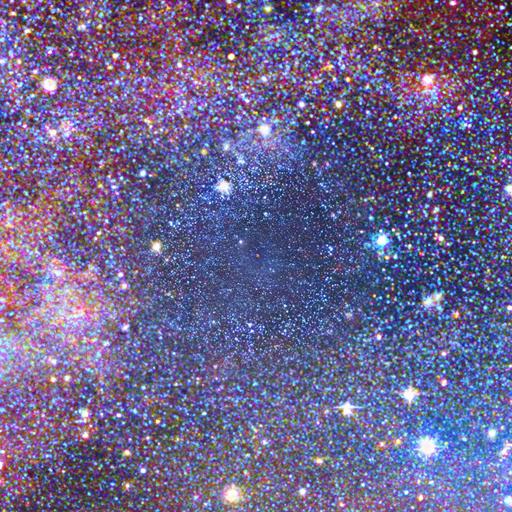} 
            &\includegraphics[width=\imW]{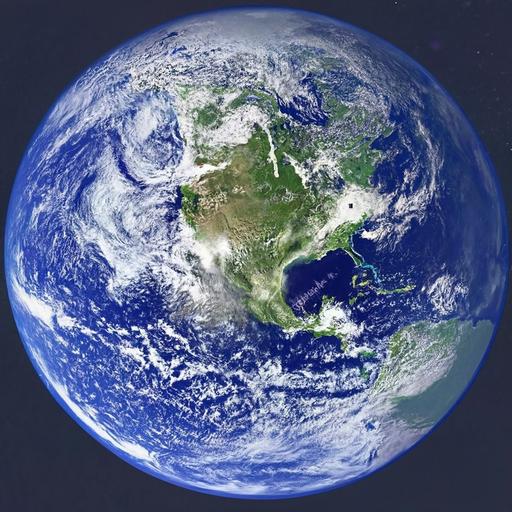} 
            &\includegraphics[width=\imW]{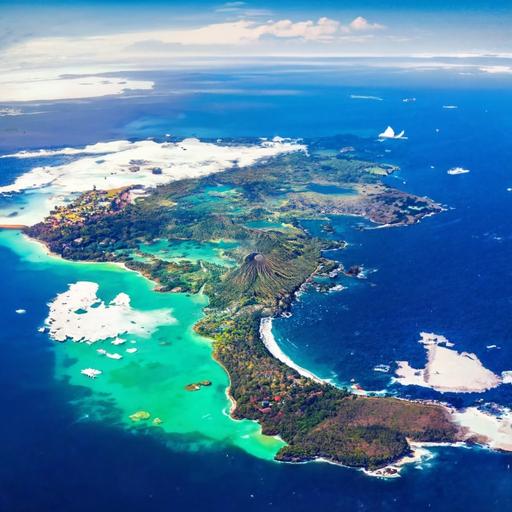} 
             &\includegraphics[width=\imW]{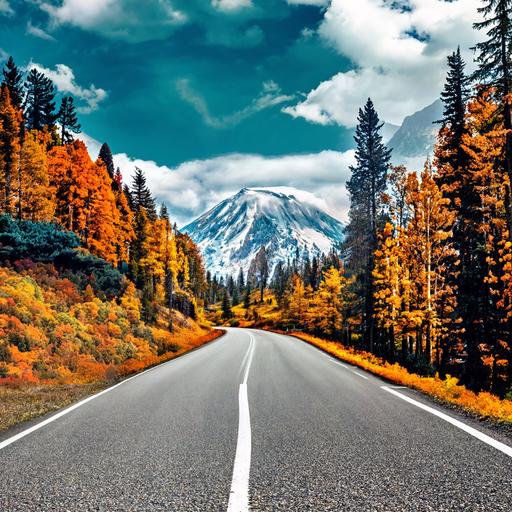} 
            \\
            \makecell{\tiny Galactic core} 
            &\makecell{\tiny Satellite image of the Earth's surface}
            &\makecell{\tiny An aerial photo capturing Hawaii's islands} 
            &\makecell{\tiny A straight road alpine forests on the sides} 
            \\
             \includegraphics[width=\imW]{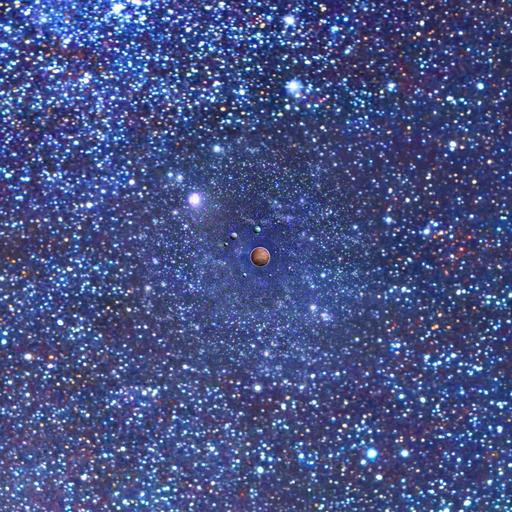} 
            &\includegraphics[width=\imW]{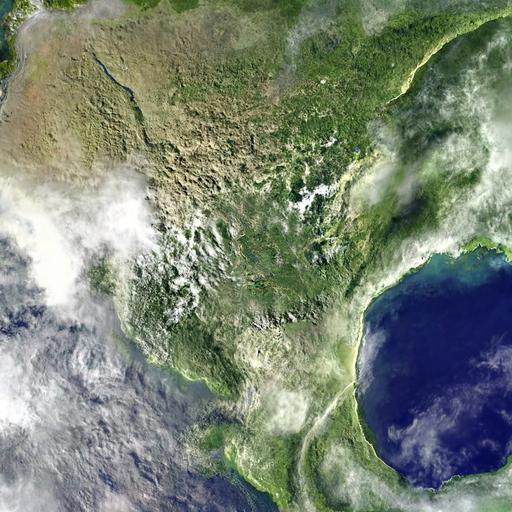} 
            &\includegraphics[width=\imW]{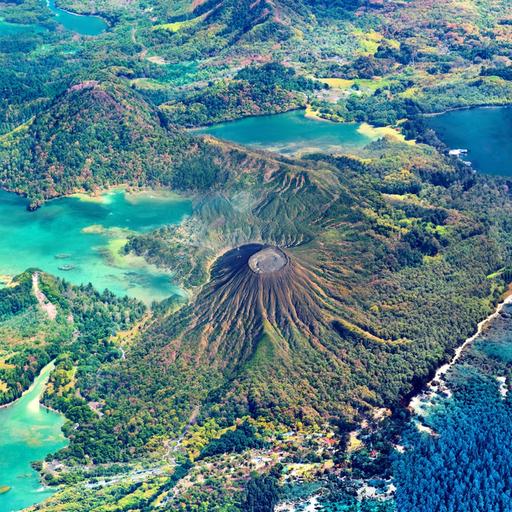} 
            &\includegraphics[width=\imW]{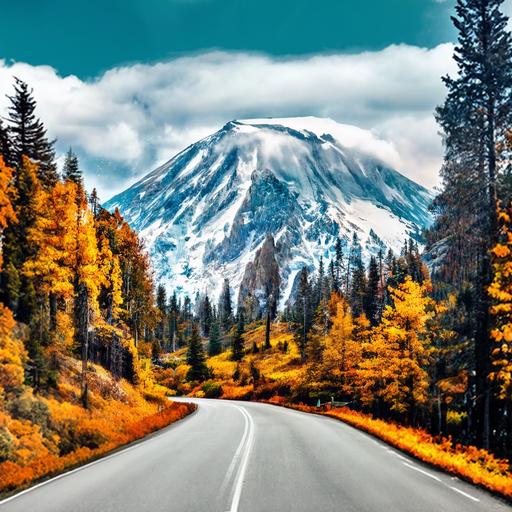} 
            \\
             \makecell{\tiny Dark starry sky} 
             &\makecell{\tiny Satellite image of a landmass of the Earth's surface} 
             &\makecell{\tiny An aerial photo of Hawaii's  mountains and rain forest} 
             &\makecell{\tiny Alpine forest road with Mount Rainier in the end} 
             \\
            \includegraphics[width=\imW]{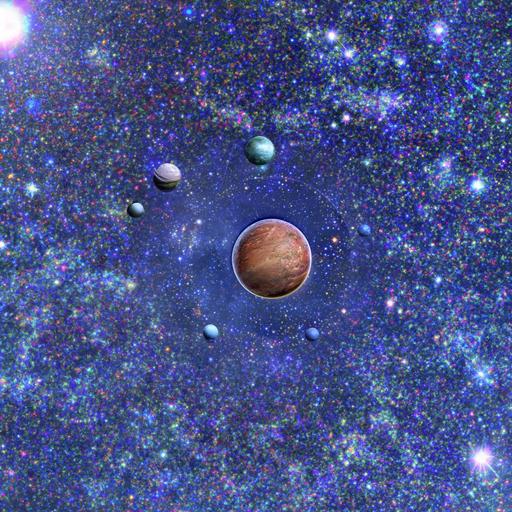} 
            &\includegraphics[width=\imW]{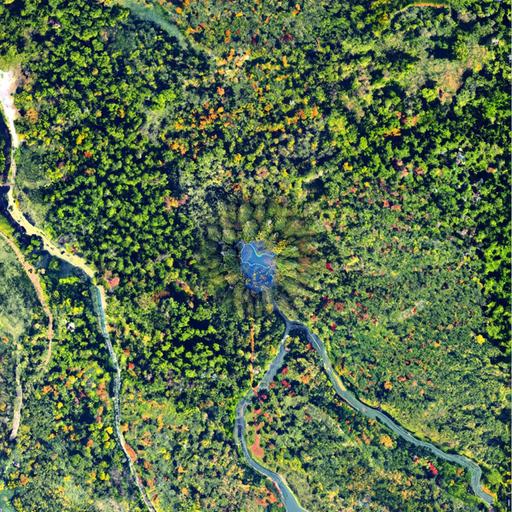} 
            &\includegraphics[width=\imW]{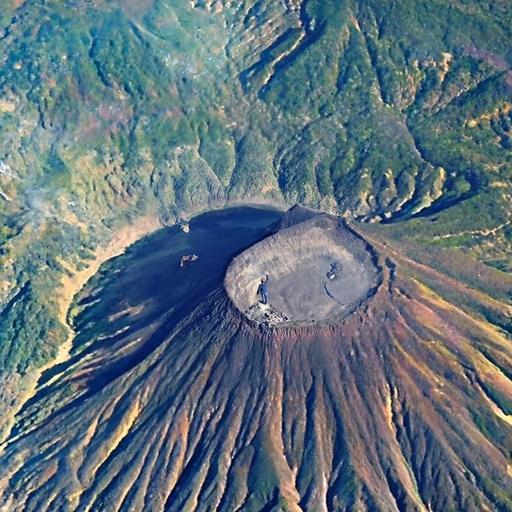} 
            & \includegraphics[width=\imW]{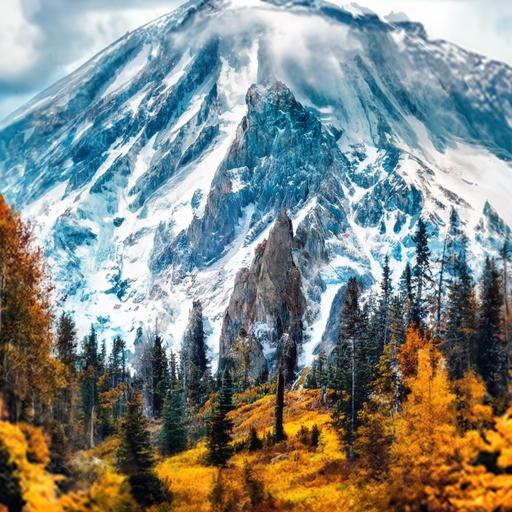}
            \\
            \makecell{\tiny Far view of alien solar system}  
            &\makecell{\tiny  Satellite image of a quaint American countryside} 
            &\makecell{\tiny An aerial close-up of the volcano's caldera} 
             &\makecell{\tiny Alpine meadows against the massive Mount Rainier} 
            \\
             \includegraphics[width=\imW]{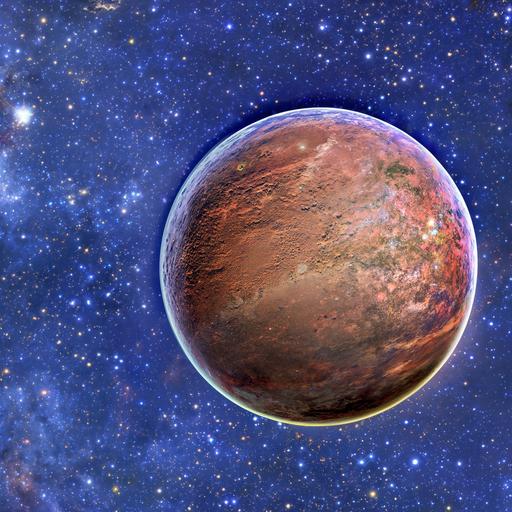} 
            &\includegraphics[width=\imW]{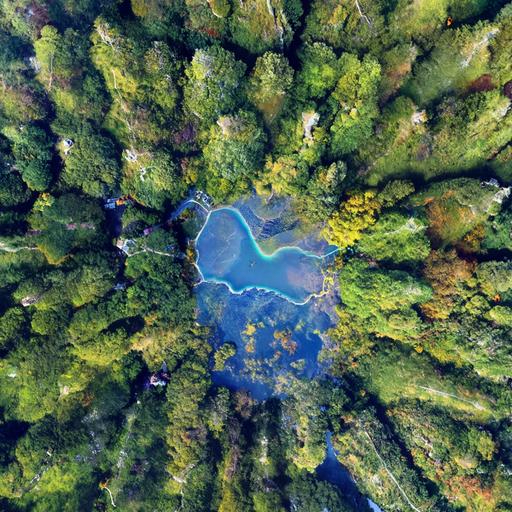} 
            &\includegraphics[width=\imW]{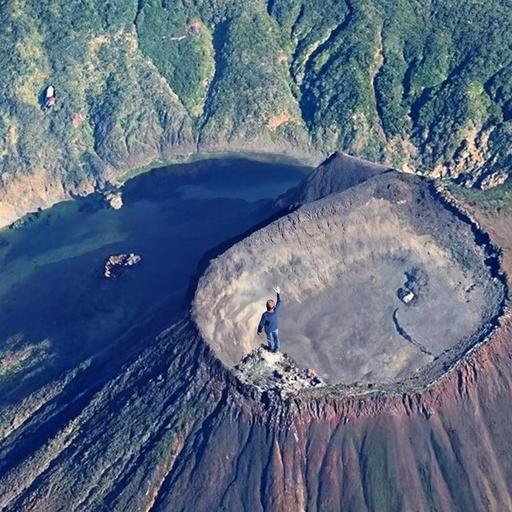} 
             &\includegraphics[width=\imW]{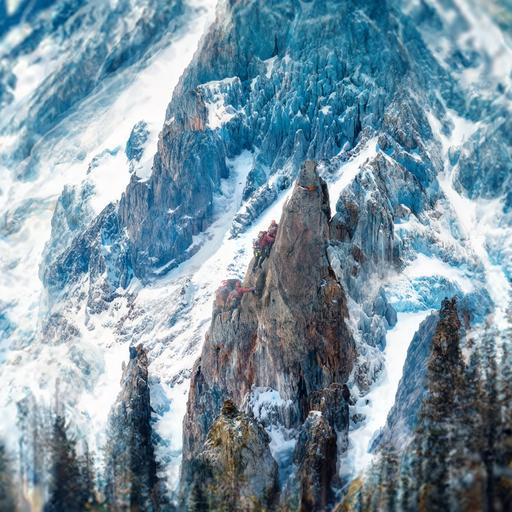} 
            \\
            \makecell{\tiny An exoplanet of a foreign solar system}  
            &\makecell{\tiny Satellite image  of a foggy forest} 
            &\makecell{\tiny An aerial close-up of the rim of a volcano's caldera} 
            &\makecell{\tiny Steep cliffs and rocky outcrops of a snow mountain } 
            \\
            \includegraphics[width=\imW]{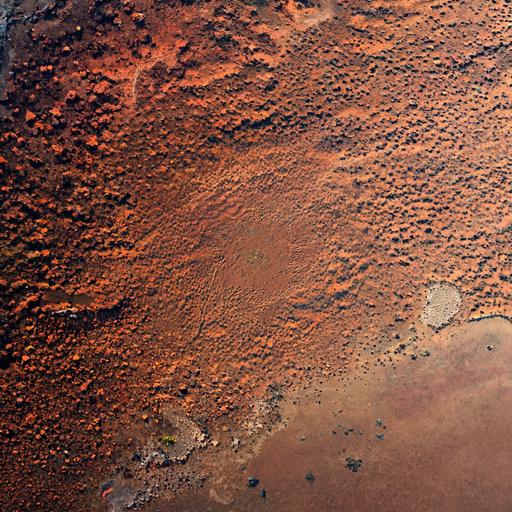} 
            &\includegraphics[width=\imW]{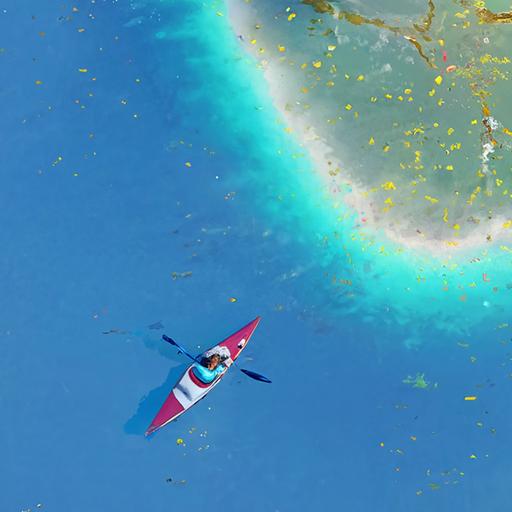} 
            &\includegraphics[width=\imW]{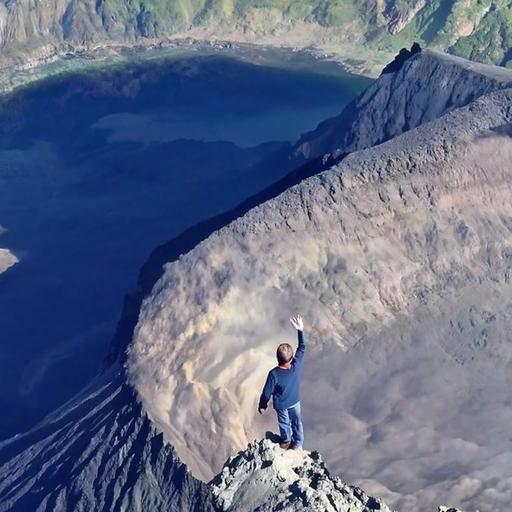} 
            &\includegraphics[width=\imW]{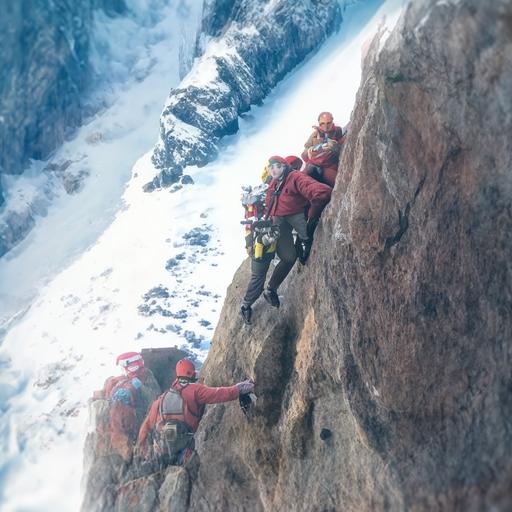} 
            \\
            \makecell{\tiny Top-down aerial image of deserted continents}  
            &\makecell{\tiny Top down view  of a lake with a person kayaking} 
            &\makecell{\tiny A man standing on the edge of a volcano's caldera} 
            &\makecell{\tiny A team of climbers climbing on the rugged cliffs} 
            \\          
        \end{tabular}
    \right\updownarrow
   \rotatebox[origin=c]{270}{\scriptsize {Zoomed out  \hspace{190mm}  Zoomed in}}$
    \caption{ Selected stills from our generated zoom videos (columns). Please refer to the supplementary materials for complete text prompts.}\label{fig:sequence_vis}
\end{figure*}

\subsection{Photograph-based Zoom}

In addition to using text prompts to generate the entire zoom stack from scratch, our approach can also generate a sequence zooming into an existing photograph. Given the most zoomed-out input image $\bm{\xi}$, we still use Alg.~\ref{alg:multiscale}, but we additionally update the denoised images to minimize the following loss function before every blending operation:
\begin{equation}
    \ell(\hat{\mathbf{x}}_{0,t}, ..., \hat{\mathbf{x}}_{N-1,t}) = \sum_{i=0}^{N-1} \|\mathcal{D}_{i}(\hat{\mathbf{x}}_{i,t})-M_i \odot \bm{\xi})\|_2^2,
\end{equation}
where, as we defined in Sec.~\ref{sec:representation}, $\mathcal{D}_i(\mathbf{x})$ downscales the image $\mathbf{x}$ by a factor $p_i$ and pads the result back to $H\times W$, and $M_i$ is a binary mask with $1$ at the center $H/{p_i}\times W/{p_i}$ square and $0$ otherwise. Before every blending operation we apply $5$ Adam~\cite{kingma2014adam} steps at a learning rate of $0.1$. This simple optimization-based strategy encourages the estimated clean images $\{\hat{\mathbf{x}}_{i,t-1}\}_{i=0}^{N-1}$ to match with the content provided in $\bm{\xi}$ in a zoom-consistent way. We show our generated photograph-based zoom sequences in Fig.~\ref{fig:real_vis}.

\subsection{Implementation Details}

For the underlying text-to-image diffusion model, we use a version of Imagen~\cite{saharia2022photorealistic} trained on internal data sources, which is a cascaded  diffusion model consisting of (1) a base model conditioned on a text prompt embedding and (2) a super resolution model additionally conditioned the low resolution output from the base model. We use its default DDPM sampling procedure with $256$ sampling steps, and we employ our {\it  multi-scale joint sampling} to the base model only. We use the super resolution model to upsample each generated image independently.
\section{Experiments}
In Figs.~\ref{fig:real_vis},~\ref{fig:sequence_vis},~\ref{fig:baseline_naive},~\ref{fig:baseline_comp1}, and~\ref{fig:ablation}, we demonstrate that our approach successfully generates consistent high quality zoom sequences for arbitrary relative zoom factors and a diverse set of scenes. Please see our supplementary materials for a full collection of videos. 
Sec.~\ref{sec:prompt} describes how we generate text prompts, Sec.~\ref{sec:baseline} demonstrates how our method outperforms diffusion-based outpainting and super-resolution models, and Sec.~\ref{sec:ablation} justifies our design decisions with an ablation study.

\subsection{Text Prompt Generation}\label{sec:prompt}
We generate a collection of text prompts that describe scenes at varying levels of scales using a combination of ChatGPT~\cite{chatgpt} and manual editing. We start with prompting ChatGPT with a description of a scene, and asking it to formulate the sequence of prompts we might need for different zoom levels. While the results from this query are often plausible, they often (1) do not accurately match the corresponding requested scales, or (2) do not match the distribution of text prompts that the text-to-image model is able to most effectively generate. As such, we manually refine the prompts. A comprehensive collection of the prompts used to generate results in the paper are provided in the supplemental materials, along with the initial versions automatically produced by ChatGPT. In the future, we expect LLMs (and in particular, multimodal models) to automatically produce a sequence of prompts well suited for this application. 
In total, we collect a total of $10$ examples, with the prompts sequence length varying form $6$ to $16$.

\subsection{Baseline Comparisons}\label{sec:baseline}
\begin{figure}[t!]
    \centering
    \def\imW{0.24\linewidth}
    \setlength{\tabcolsep}{1pt}
    \renewcommand{\arraystretch}{0.5}
    \begin{tabular}{cccc}
      \multicolumn{4}{c}{$\xlongleftrightarrow{\small{\text{Zoomed out} \hspace{55mm} \text{Zoomed in}}}$}\\
        \includegraphics[width=\imW]{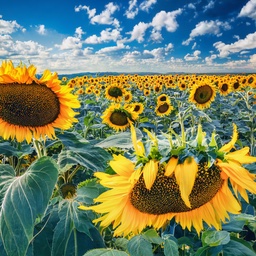}
        &\includegraphics[width=\imW]{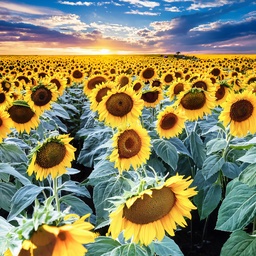}
        &\includegraphics[width=\imW]{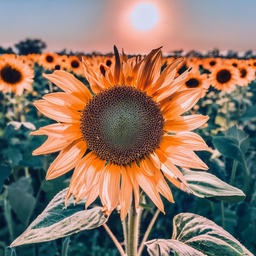}
        &\includegraphics[width=\imW]{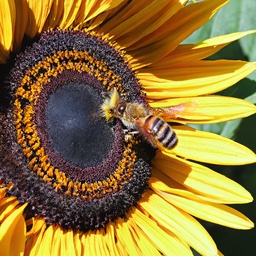}\\
        \includegraphics[width=\imW]{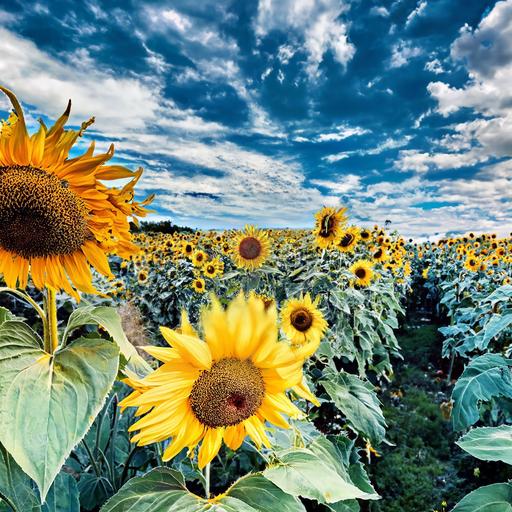}
        &\includegraphics[width=\imW]{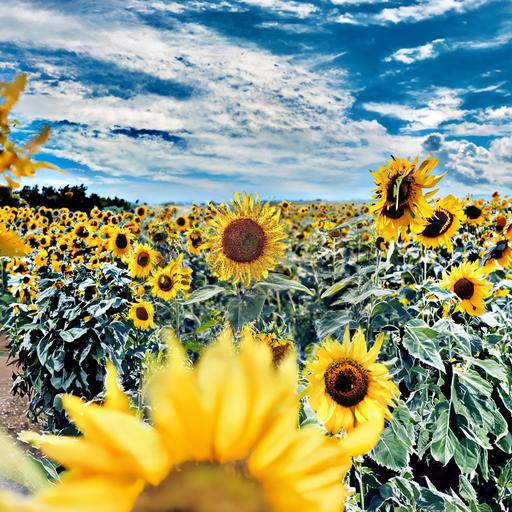}
        &\includegraphics[width=\imW]{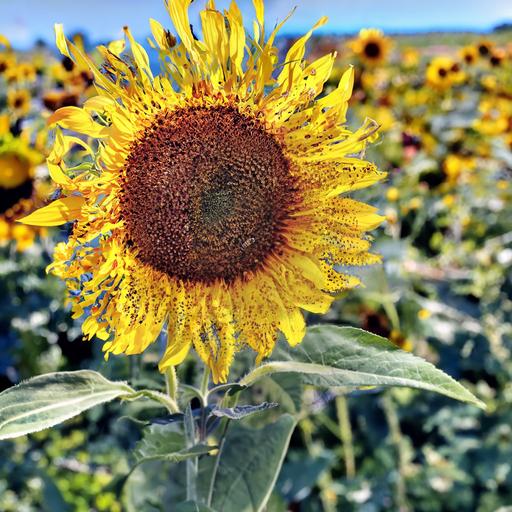}
        &\includegraphics[width=\imW]{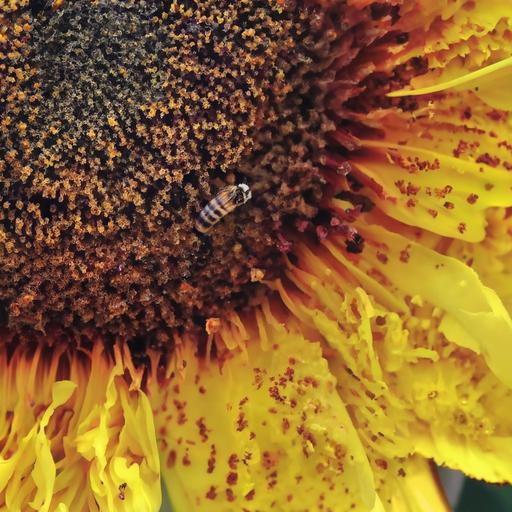}
    \end{tabular}
    \caption{Generated zoom sequences with independent sampling (top) and our multi-scale sampling (bottom). Our method encourages different levels to depict a consistent underlying scene, while not compromising the image quality. }\label{fig:baseline_naive}
    \vspace{-10pt}
\end{figure}

Fig.~\ref{fig:baseline_naive} compares zoom sequences generated with our method and without (\ie, independently sampling each scale). When compared to our results, the independently-generated images similarly follow the text prompt, but clearly do not correspond to a single consistent underlying scene. 

\begin{figure*}[ht!]
    \centering
    \def\imW{0.15\linewidth}
    \setlength{\tabcolsep}{1pt}
     \renewcommand\cellset{\renewcommand\arraystretch{0}}%
    \renewcommand{\arraystretch}{0.5}
    $\left.
    \begin{tabular}{ccc@{\hskip 8pt} ccc}
       {\small SR} & {\small Outpainting} & {\small Ours}  &{\small SR} & {\small Outpainting} & {\small Ours} \\
        \includegraphics[width=\imW, height=\imW]{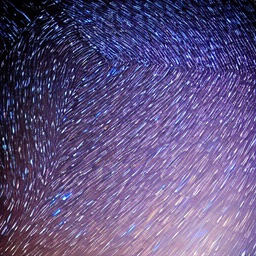}
        &\includegraphics[width=\imW, height=\imW]{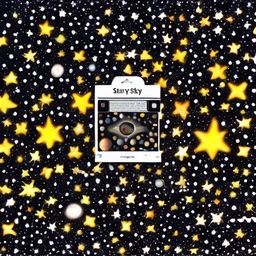}
        &\includegraphics[width=\imW, height=\imW]{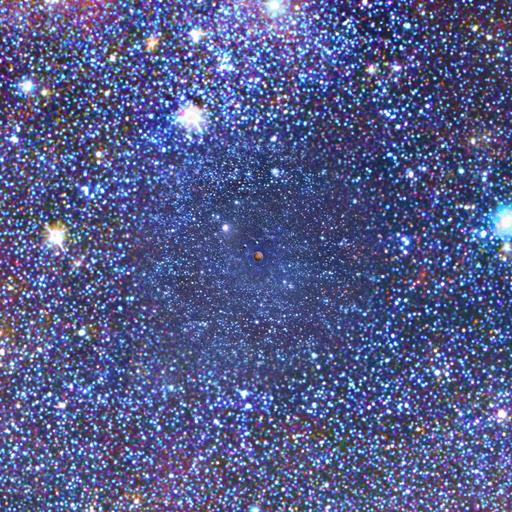} 
        &\includegraphics[width=\imW, height=\imW]{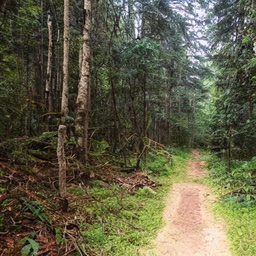}
        &\includegraphics[width=\imW, height=\imW]{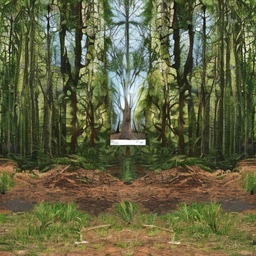}
        &\includegraphics[width=\imW, height=\imW]{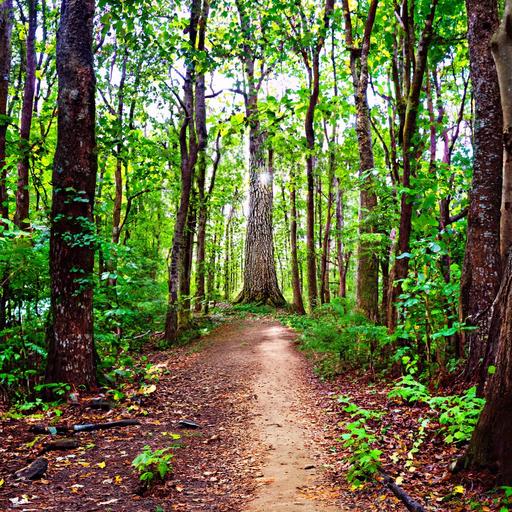}\\
         \multicolumn{3}{c}{\scriptsize Thousands of stars against dark space in the background}
        &\multicolumn{3}{c}{\scriptsize  Path leading to the dense forest from open land} \\
         \includegraphics[width=\imW, height=\imW]{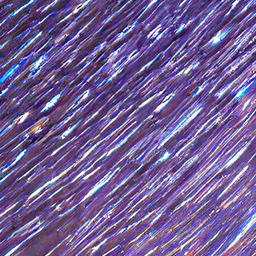}
        &\includegraphics[width=\imW, height=\imW]{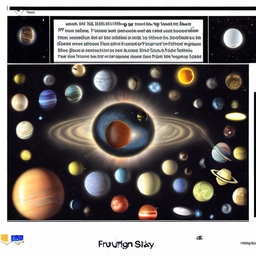}
        &\includegraphics[width=\imW, height=\imW]{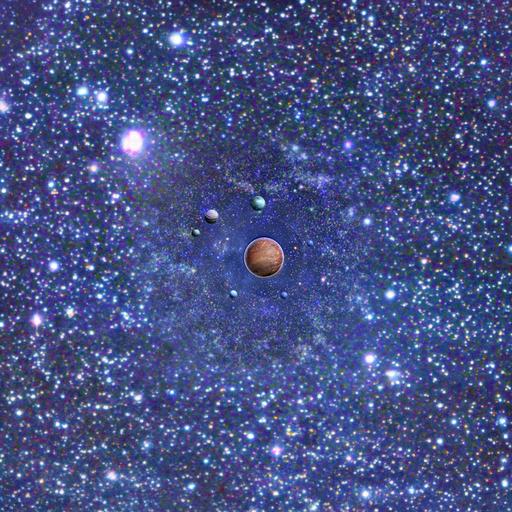} 
        &\includegraphics[width=\imW, height=\imW]{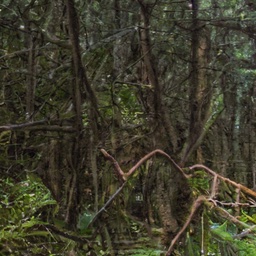}
        &\includegraphics[width=\imW, height=\imW]{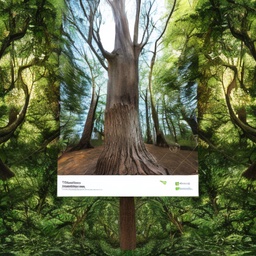} 
        &\includegraphics[width=\imW, height=\imW]{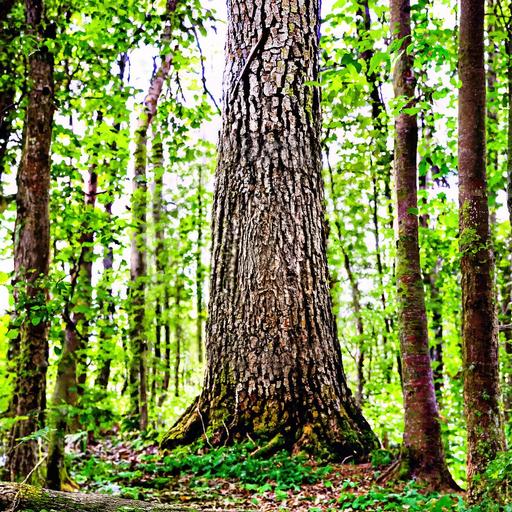} \\
         \multicolumn{3}{c}{\scriptsize Dark starry sky with a foreign solar system in the middle} 
       & \multicolumn{3}{c}{\scriptsize Heart of a forest filled with tree trunks, leaves, vines, and undergrowth} \\
          \includegraphics[width=\imW, height=\imW]{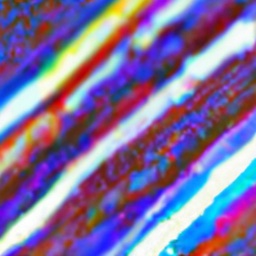}
        &\includegraphics[width=\imW, height=\imW]{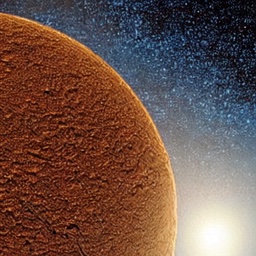}
         &\includegraphics[width=\imW, height=\imW]{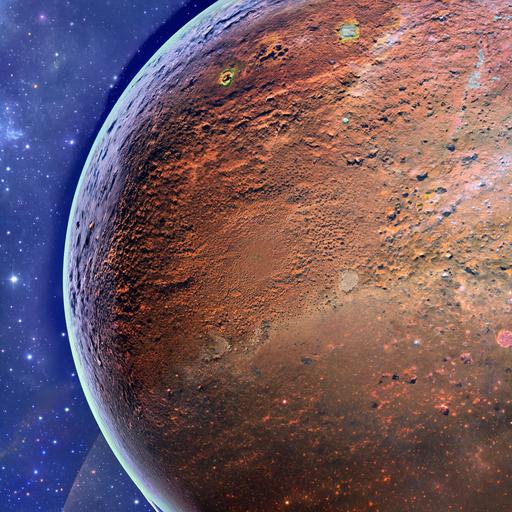} 
        & \includegraphics[width=\imW, height=\imW]{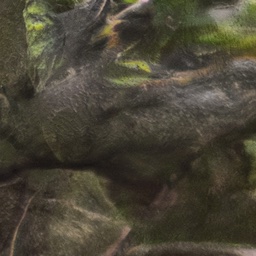}
        &\includegraphics[width=\imW, height=\imW]{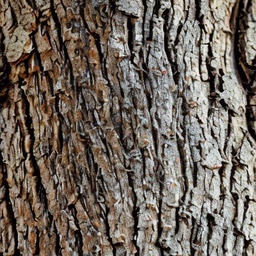}
        &\includegraphics[width=\imW, height=\imW]{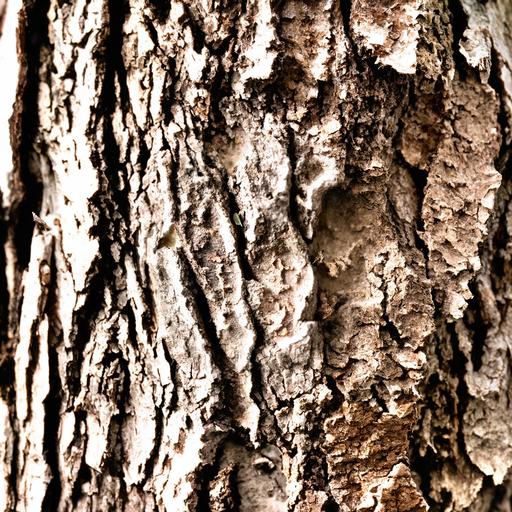} \\
        \multicolumn{3}{c}{\scriptsize A close-up of an exoplanet in a foreign solar system} 
         &\multicolumn{3}{c}{\scriptsize Detailed view of an oak tree bark showing ridges and grooves} \\
    \end{tabular}
    \right\updownarrow
    \rotatebox[origin=c]{270}{\scriptsize{Zoomed out  \hspace{65mm}  Zoomed in}}$
    \caption{Comparisons with Stable Diffusion Outpainting and super-resolution (SR) models.}\label{fig:baseline_comp1}
\end{figure*}

Next, we compare our method to two autogressive generation approaches for generating zoom sequences: (1) Stable Diffusion's~\cite{stabilityai} outpainting model and (2) Stable Diffusion's ``upscale'' super-resolution model. 
We show representative qualitative results in Fig.~\ref{fig:baseline_comp1}. 

\vspace{0.1cm}
{\bf \noindent Comparison to progressive outpainting.} 
The outpainting baseline starts with generating the most zoomed-in image and progressively generates coarser scales by downsampling the previous generated image and outpainting the surrounding area. As in our method, the inpainting of each level is conditioned on the corresponding text prompt. In Fig.~\ref{fig:baseline_comp1}, we show that because of the causality of the autoregressive process, the outpainting approach suffers from gradually accumulating errors, \ie, when a mistake is made at a given step, later outpainting iterations may struggle to produce a consistent image.

\vspace{0.1cm}
{\bf \noindent Comparison to progressive super-resolution.} 
The super-resolution baseline starts with the most zoomed-out image and generates subsequent scales by super-resolving the upscaled central image region, conditioned on the corresponding text prompt. The low resolution input provides strong structural information which constrains the layout of the next zoomed-in image. As we can see in Fig.~\ref{fig:baseline_comp1}, this super-resolution baseline is not able to synthesize new objects that would only appear in the finer, zoomed-in scales.

\subsection{Ablations}\label{sec:ablation}
\begin{figure}[h]
    \centering
    \def\imW{0.22\linewidth}
    \setlength{\tabcolsep}{1pt}
    \renewcommand\cellset{\renewcommand\arraystretch{0.5}}%
    \renewcommand{\arraystretch}{0.5}
    $\left.
    \begin{tabular}{cccc}
        \makecell{{\scriptsize Iterative update}} &\makecell{{\scriptsize w/o Shared noise}}   & \makecell{ {\scriptsize Naïve blending}} & \makecell{\scriptsize  Ours} \\
        \includegraphics[width=\imW, height=\imW]{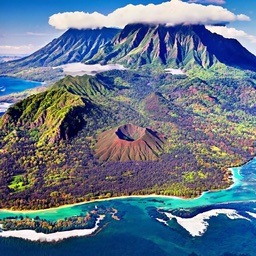} 
        &\includegraphics[width=\imW, height=\imW]{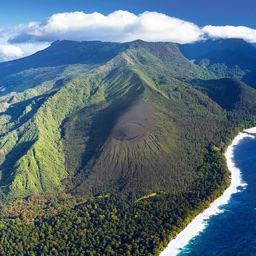}
        &\includegraphics[width=\imW, height=\imW]{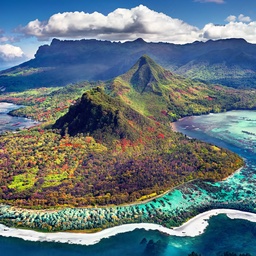} 
         &\includegraphics[width=\imW, height=\imW]{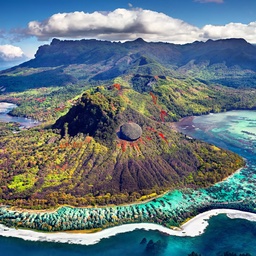}
         \\
       \includegraphics[width=\imW, height=\imW]{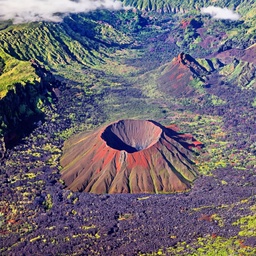} 
        &\includegraphics[width=\imW, height=\imW]{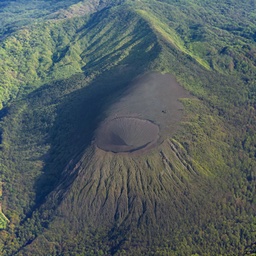}
        &\includegraphics[width=\imW, height=\imW]{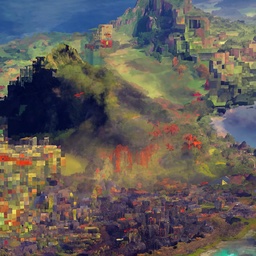} 
         &\includegraphics[width=\imW, height=\imW]{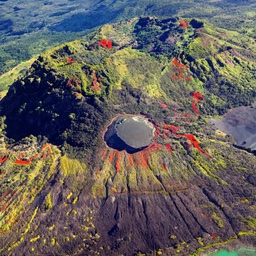}
         \\
       \includegraphics[width=\imW, height=\imW]{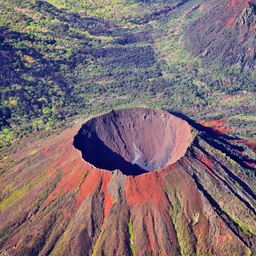} 
        &\includegraphics[width=\imW, height=\imW]{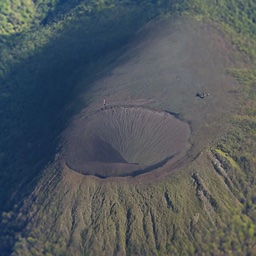}
        &\includegraphics[width=\imW, height=\imW]{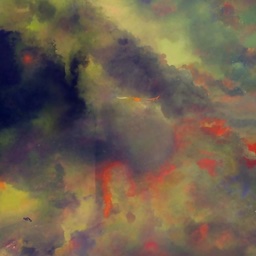} 
         &\includegraphics[width=\imW, height=\imW]{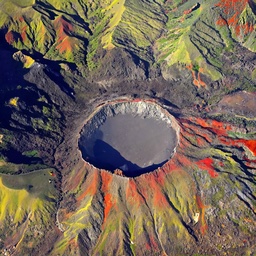}
         \\
      \includegraphics[width=\imW, height=\imW]{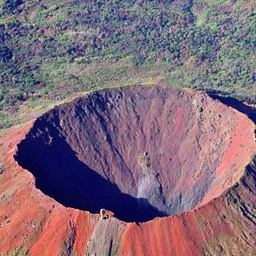} 
      &\includegraphics[width=\imW, height=\imW]{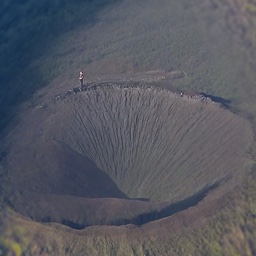}
        &\includegraphics[width=\imW, height=\imW]{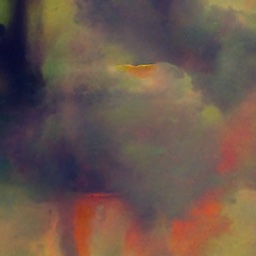}
         &\includegraphics[width=\imW, height=\imW]{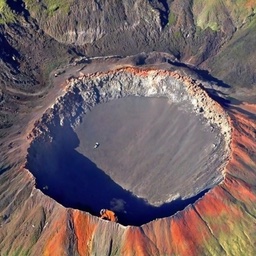}
         \\
    \end{tabular}
    \right\updownarrow
   \rotatebox[origin=c]{270}{\scriptsize {Zoomed out  \hspace{55mm}  Zoomed in}}$
    \caption{ {\bf Ablations.} We evaluate other options for multi-scale consistency: (1) iteratively updating each level separately, (2) na\"ive multi-scale blending, (3) removing the shared noise.}\label{fig:ablation}
    \vspace{-10pt}
\end{figure}

In Fig.~\ref{fig:ablation}, we show comparisons to simpler versions of our method to examine the effect of our design decisions.

\vspace{0.1cm}
{\bf \noindent Joint vs. Iterative update.} 
Instead of performing multi-scale blending approach, we can instead iteratively cycle through the images in the zoom stack, and perform one sampling step at each level independently. Unlike fully independent sampling, this process does allow for sharing of information between scales, since the steps are still applied to renders from the zoom stack. We find that although this produces more consistent results than independent sampling, there remain inconsistencies at stack layer boundaries.

\vspace{0.1cm}  
{\bf \noindent Shared vs. random noise }
Instead of using a shared noise $\Pi_{\text{noise}}$, noise can be sampled independently for each zoom level. We find that this leads to blur in the output samples. 

\vspace{0.1cm}
{\bf \noindent Comparison with naïve blending.}
Instead of our multi-scale blending, we can instead na\"ively blend the observations together, \eg, as in MultiDiffusion~\cite{bar2023multidiffusion}. We find that this leads to blurry outputs at deeper zoom levels.

\section{Discussion \& Limitations}
A significant challenge in our work is discovering the appropriate set of text prompts that (1) agree with each other across a set of fixed scales, and (2) can be effectively generated consistently by a given text-to-image model. One possible avenue of improvement could be to, along with sampling, optimize for suitable geometric transformations between successive zoom levels. These transformations could include translation, rotation, and even scale, to find better alignment between the zoom levels and the prompts.

Alternatively, one can optimize the text embeddings, to find better descriptions that correspond to subsequent zoom levels. Or, instead, use the LLM for in-the-loop generation, \ie, by giving LLM the generated image content, and asking it to refine its prompts to produce images which are closer in correspondence given the set of pre-defined scales.

\vspace{0.1cm}
{\bf \noindent Acknowledgements.} We thank Ben Poole, Jon Barron, Luyang Zhu, Ruiqi Gao, Tong He, Grace Luo, Angjoo Kanazawa, Vickie Ye, Songwei Ge, Keunhong Park, and David Salesin for helpful discussions and feedback. This work was supported in part by UW Reality Lab, Meta, Google, OPPO, and Amazon.
{
    \small
    \bibliographystyle{ieeenat_fullname}
    \bibliography{main}
}
\appendix
\setcounter{page}{1}

\noindent \underline{\textbf{Please see our supp. video for a full set of video results.}}

\section{Additional comparisons} 
\begin{figure}[h]
    \centering
    \def\imW{0.24\linewidth}
    \setlength{\tabcolsep}{1pt}
     \renewcommand\cellset{\renewcommand\arraystretch{0}}%
    \renewcommand{\arraystretch}{0.5}
    \begin{tabular}{cccc}
     \multicolumn{4}{c}{$\xlongleftrightarrow{\small{\text{Zoomed out} \hspace{55mm} \text{Zoomed in}}}$}\\
     \includegraphics[width=\imW, height=\imW]{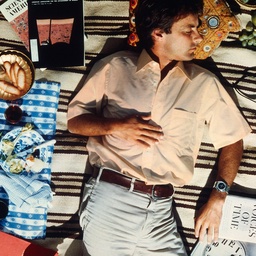}
     &\includegraphics[width=\imW, height=\imW]{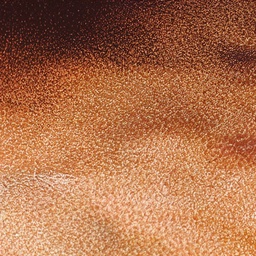}
     &\includegraphics[width=\imW, height=\imW]{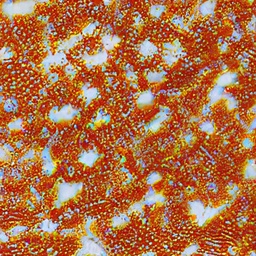}
     &\includegraphics[width=\imW, height=\imW]{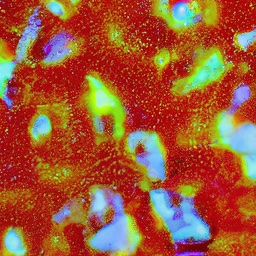}\\
      \includegraphics[width=\imW, height=\imW]{figs/ZoominSequences/Hand/00.jpg}
     &\includegraphics[width=\imW, height=\imW]{figs/ZoominSequences/Hand/02.jpg}
     &\includegraphics[width=\imW, height=\imW]{figs/ZoominSequences/Hand/04.jpg}
     &\includegraphics[width=\imW, height=\imW]{figs/ZoominSequences/Hand/06.jpg}\\
    \end{tabular}
    \caption{Comparison between the Stable Diffusion super-resolution model (top) and our method (bottom), zooming into a scene defined by a provided real input image (left).}\label{fig:baseline_comp_sr}
\end{figure}

\begin{figure*}
    \centering
    \def\imW{0.15\linewidth}
    \setlength{\tabcolsep}{1pt}
     \renewcommand\cellset{\renewcommand\arraystretch{0}}%
    \renewcommand{\arraystretch}{0.5}
    \begin{tabular}{ccc@{\hskip 8pt} ccc}
       {\small SR} & {\small Outpainting} & {\small Ours}  &{\small SR} & {\small Outpainting} & {\small Ours} \\
        \includegraphics[width=\imW, height=\imW]{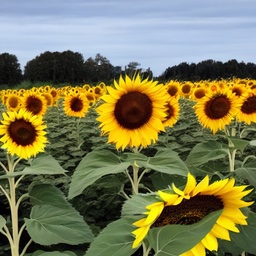}
        &\includegraphics[width=\imW, height=\imW]{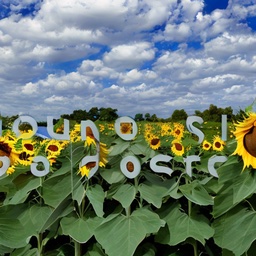}
        &\includegraphics[width=\imW, height=\imW]{figs/ZoominSequences/Sunflowers/00.jpg} 
        &\includegraphics[width=\imW, height=\imW]{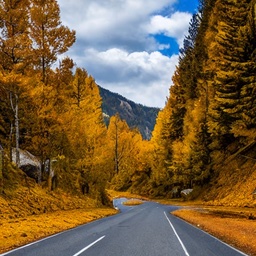}
        &\includegraphics[width=\imW, height=\imW]{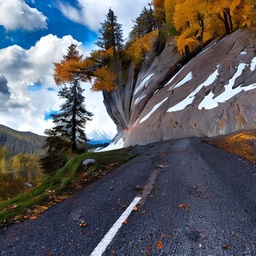}
        &\includegraphics[width=\imW, height=\imW]{figs/ZoominSequences/Rainier/00.jpg}\\
         \multicolumn{3}{c}{\scriptsize A sunflower field from afar}
        &\multicolumn{3}{c}{\scriptsize  A straight road alpine forests on the sides} \\
         \includegraphics[width=\imW, height=\imW]{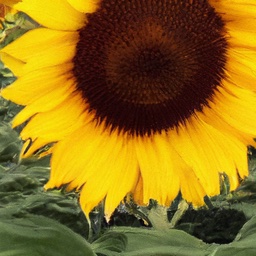}
        &\includegraphics[width=\imW, height=\imW]{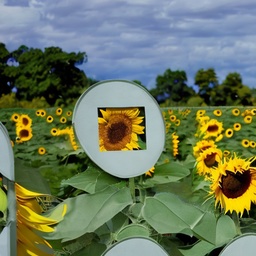}
        &\includegraphics[width=\imW, height=\imW]{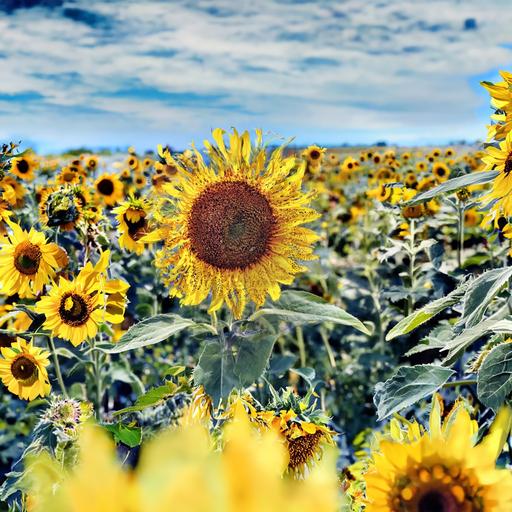} 
        &\includegraphics[width=\imW, height=\imW]{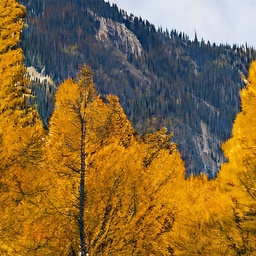}
        &\includegraphics[width=\imW, height=\imW]{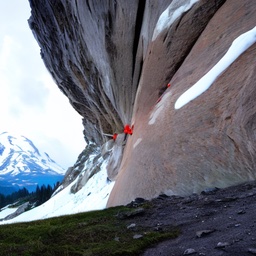} 
        &\includegraphics[width=\imW, height=\imW]{figs/ZoominSequences/Rainier/02.jpg} \\
         \multicolumn{3}{c}{\scriptsize Close-up of rows of sunflowers} 
       & \multicolumn{3}{c}{\scriptsize Alpine meadows against the massive Mount Rainier} \\
        \includegraphics[width=\imW, height=\imW]{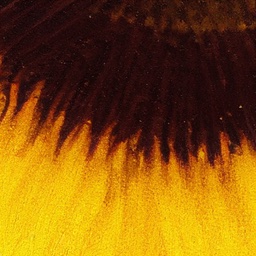}
        &\includegraphics[width=\imW, height=\imW]{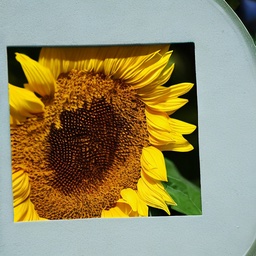}
         &\includegraphics[width=\imW, height=\imW]{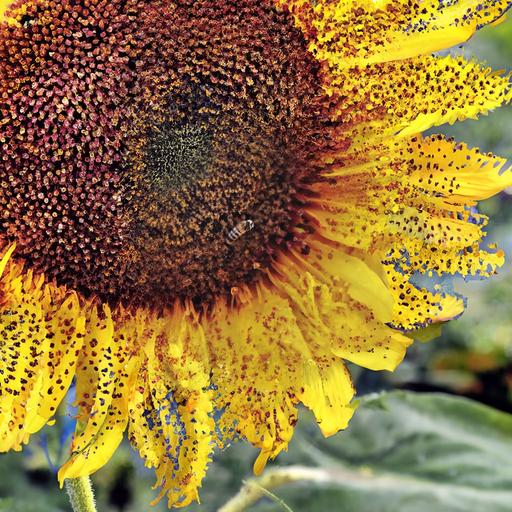} 
        & \includegraphics[width=\imW, height=\imW]{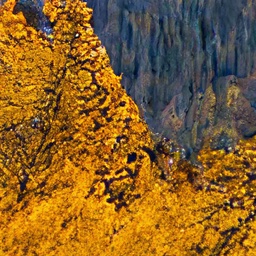}
        &\includegraphics[width=\imW, height=\imW]{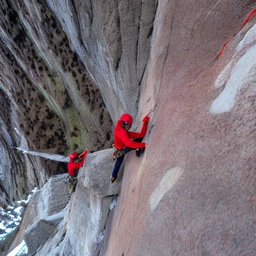}
        &\includegraphics[width=\imW, height=\imW]{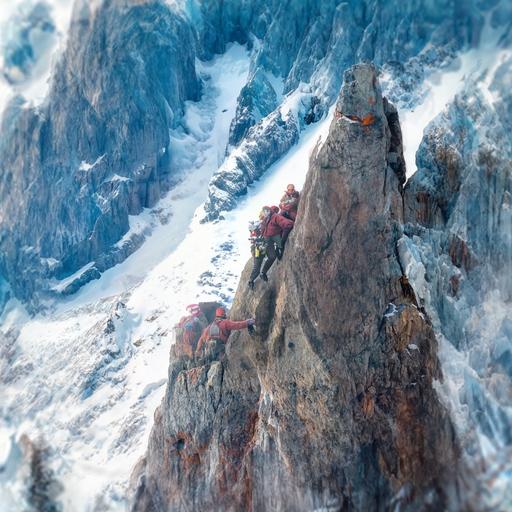} \\
        \multicolumn{3}{c}{\scriptsize A closer view of the sunflower in the center with its golden petals} 
         &\multicolumn{3}{c}{\scriptsize Steep cliffs and rocky outcrops of a snow mountain} \\
     \midrule
     
      \includegraphics[width=\imW, height=\imW]{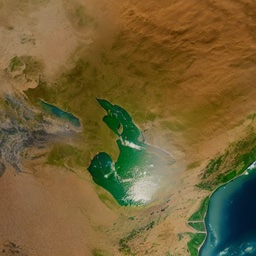}
        &\includegraphics[width=\imW, height=\imW]{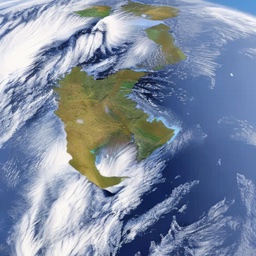}
        &\includegraphics[width=\imW, height=\imW]{figs/ZoominSequences/Earth/00.jpg} 
        &\includegraphics[width=\imW, height=\imW]{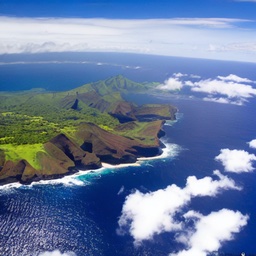}
        &\includegraphics[width=\imW, height=\imW]{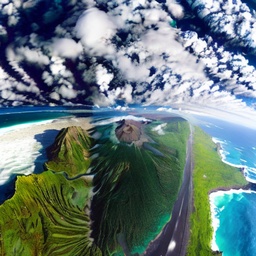}
        &\includegraphics[width=\imW, height=\imW]{figs/ZoominSequences/Hawaii/00.jpg}\\
         \multicolumn{3}{c}{\scriptsize Satellite image of the Earth's surface}
        &\multicolumn{3}{c}{\scriptsize An aerial photo capturing Hawaii's islands} \\
         \includegraphics[width=\imW, height=\imW]{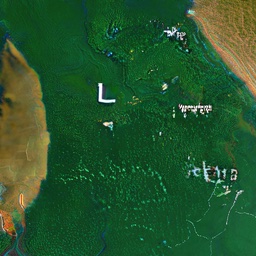}
        &\includegraphics[width=\imW, height=\imW]{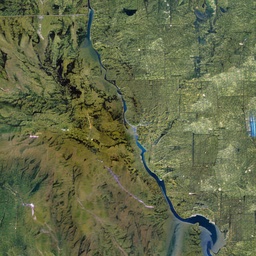}
        &\includegraphics[width=\imW, height=\imW]{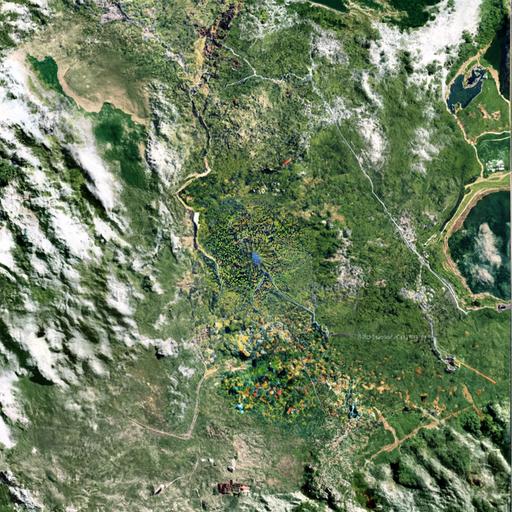} 
        &\includegraphics[width=\imW, height=\imW]{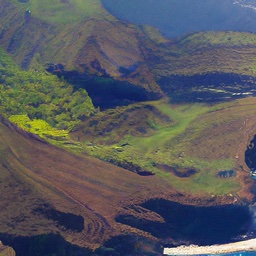}
        &\includegraphics[width=\imW, height=\imW]{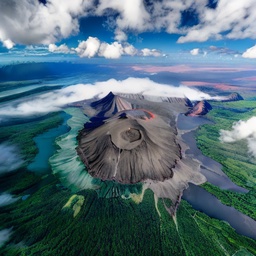} 
        &\includegraphics[width=\imW, height=\imW]{figs/ZoominSequences/Hawaii/02.jpg} \\
         \multicolumn{3}{c}{\scriptsize Satellite image of a state in the U.S., with its natural beauty} 
       & \multicolumn{3}{c}{\scriptsize An aerial photo of Hawaii's  mountains and rain forest} \\
        \includegraphics[width=\imW, height=\imW]{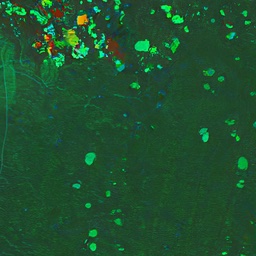}
        &\includegraphics[width=\imW, height=\imW]{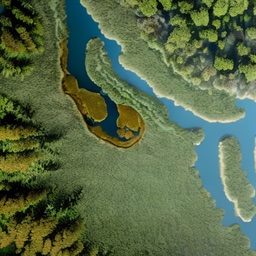}
         &\includegraphics[width=\imW, height=\imW]{figs/ZoominSequences/Earth/04.jpg} 
        & \includegraphics[width=\imW, height=\imW]{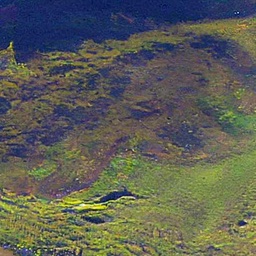}
        &\includegraphics[width=\imW, height=\imW]{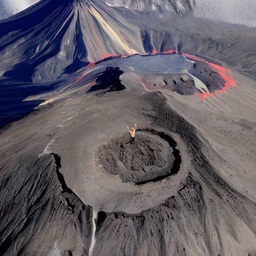}
        &\includegraphics[width=\imW, height=\imW]{figs/ZoominSequences/Hawaii/04.jpg} \\
        \multicolumn{3}{c}{\scriptsize Satellite image  of a foggy forest with a lake in the middle} 
         &\multicolumn{3}{c}{\scriptsize  An aerial close-up of the volcano's caldera} \\
         \includegraphics[width=\imW, height=\imW]{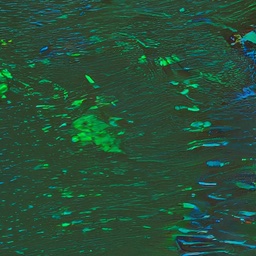}
        &\includegraphics[width=\imW, height=\imW]{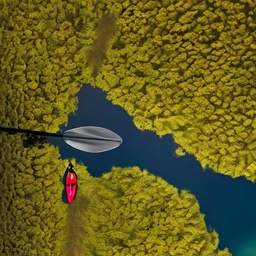}
         &\includegraphics[width=\imW, height=\imW]{figs/ZoominSequences/Earth/06.jpg} 
        & \includegraphics[width=\imW, height=\imW]{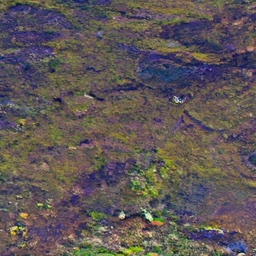}
        &\includegraphics[width=\imW, height=\imW]{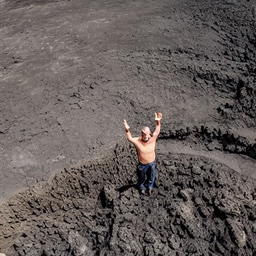}
        &\includegraphics[width=\imW, height=\imW]{figs/ZoominSequences/Hawaii/06.jpg} \\
        \multicolumn{3}{c}{\scriptsize Top down view  of a lake with a person kayaking} 
         &\multicolumn{3}{c}{\scriptsize A man standing on the edge of a volcano's caldera} \\
    \end{tabular}
    \caption{Comparisons with Stable Diffusion Outpainting and super-resolution (SR) models.}\label{fig:baseline_comp2}
    \vspace{-10pt}
\end{figure*}

We show additional qualitative comparisons with super resolution and outpainting models in Fig.~\ref{fig:baseline_comp2}. In Fig.~\ref{fig:baseline_comp_sr}, we compare with the super resolution model for photograph-based zoom.

\section{Quantitative evaluations}
We conduct a user study involving $38$ participants who were presented with a set of $18$ pairwise comparisons of our method and one of the these two baselines. Participants were asked to select one of the two options in response to the question, ``Which [...] looks like a camera zooming into a consistent scene?"---our method was chosen in $92\%$ of $684$ responses. 

In addition, we report (1) CLIP scores which measure text-image alignment, and (2) CLIP aesthetic scores (from MultiDiffusion~\cite{bar2023multidiffusion}), which measure image aesthetic quality on the generated images, using our method and baseline methods. The scores are shown in Tab.~\ref{tab:quant}.

\begin{table}[h]
    \centering
    \begin{tabular}{c|ccc}
         &{\small CLIP-score $\uparrow$} &{\small CLIP-aesthetic $\uparrow$}  \\
    \hline
    {\small SR} 
    &29.18 &4.89 \\
    {\small Outpainting} 
    &30.08 &5.51 \\
    {\small Ours} 
    &{\bf 31.39} &{\bf 5.65}
    \end{tabular}
    \caption{Quantitative evaluation compared with baselines. Metrics computed at all prompt scales and averaged across all examples.}
    \label{tab:quant}
\end{table}

\section{Text prompts generation}
\label{sec:prmopts}
As mentioned in the main paper, large language models are a viable option for generating text prompts that describe a scene at various zoom levels, but their outputs are often imperfect---either describing scales that do not perfectly correspond to the scale factors used in sampling, or describing content with text phrases that do not match the learned distribution of the text-to-image model. In these cases, we often find it necessary to make manual adjustments to the set of text prompts. We show a comparison of the prompts generated by ChatGPT and the corresponding manually refined prompts (which were used to generate our zooming videos) in Tab.~\ref{tab:chatgpt}. Some sequences were not generated automatically---these are shown in Tabs.~\ref{tab:rainier}, ~\ref{tab:galaxy},~\ref{tab:hand}, and~\ref{tab:leaf}.

\section{Effect of prompts}
\begin{figure}[h]
    \def\imW{0.16\linewidth}
    \setlength{\tabcolsep}{0.8pt}
    \renewcommand{\arraystretch}{0.5}
    \begin{tabular}{cccccc}
        \includegraphics[width=\imW]{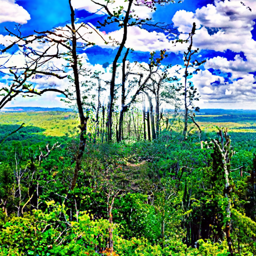}
        &\includegraphics[width=\imW]{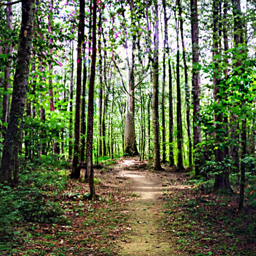}
        &\includegraphics[width=\imW]{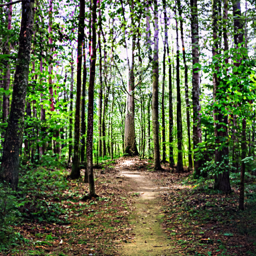}
        &\includegraphics[width=\imW]{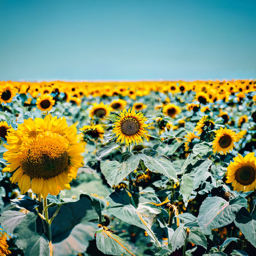}
        &\includegraphics[width=\imW]{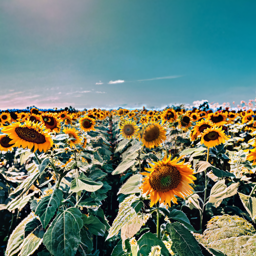}
        &\includegraphics[width=\imW]{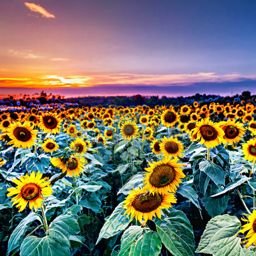} \\
        
        \includegraphics[width=\imW, align=c]{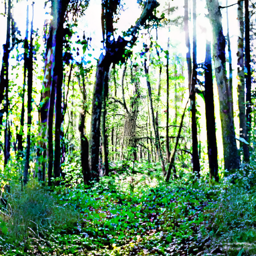}
        &\includegraphics[width=\imW, align=c]{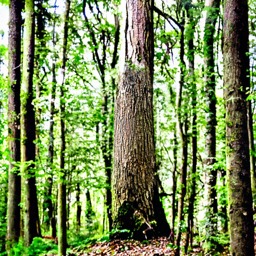}
        &\includegraphics[width=\imW, align=c]{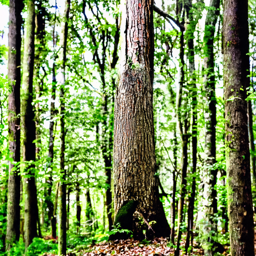}
        & \makecell[c]{{\scriptsize level added} \\ {\scriptsize in refinement}}
        &\includegraphics[width=\imW, align=c]{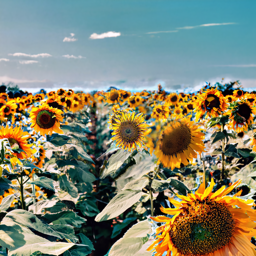}
        &\includegraphics[width=\imW, align=c]{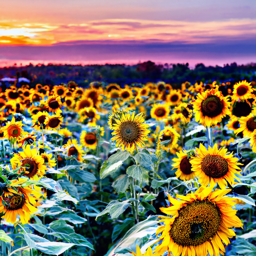} \\

        \includegraphics[width=\imW]{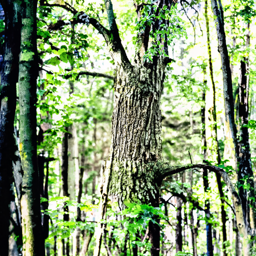}
        &\includegraphics[width=\imW]{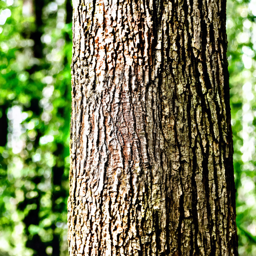}
        &\includegraphics[width=\imW]{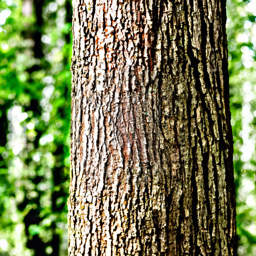}
        &\includegraphics[width=\imW]{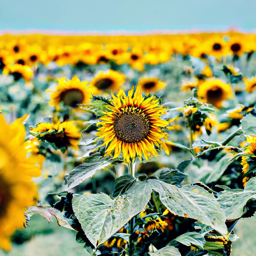}
        &\includegraphics[width=\imW]{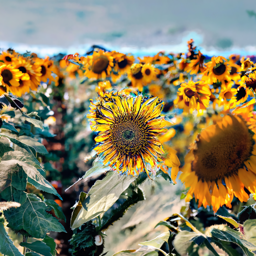}
        &\includegraphics[width=\imW]{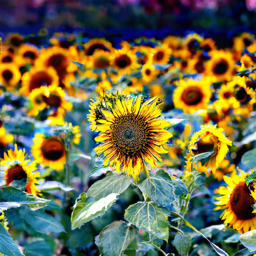}\\
         
        \includegraphics[width=\imW]{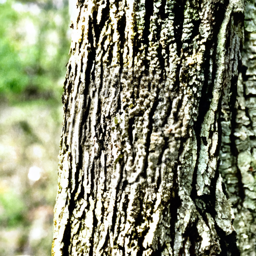}
        &\includegraphics[width=\imW]{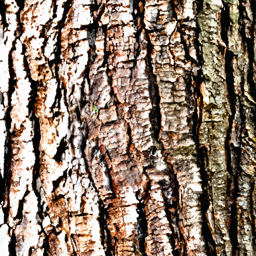}
        &\includegraphics[width=\imW]{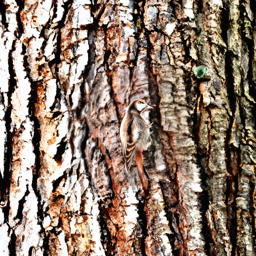}
        &\includegraphics[width=\imW]{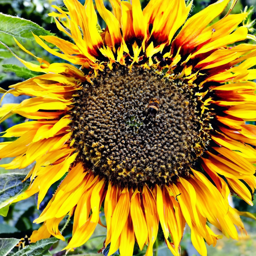}
        &\includegraphics[width=\imW]{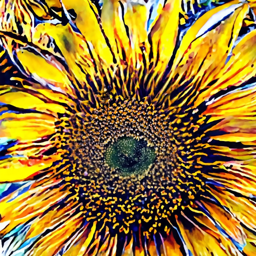}
        &\includegraphics[width=\imW]{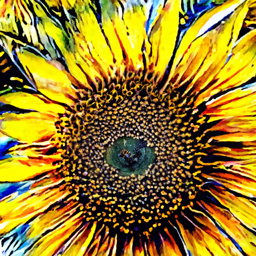}\\
        
        \includegraphics[width=\imW]{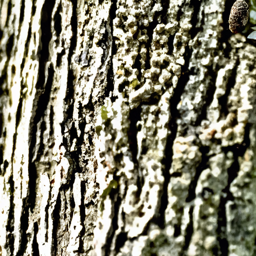}
        &\includegraphics[width=\imW]{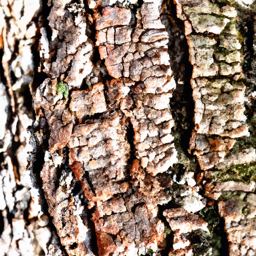}
        &\includegraphics[width=\imW]{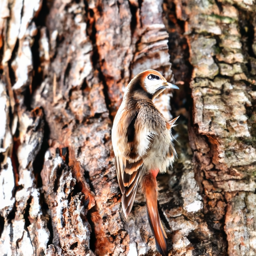}
        &\includegraphics[width=\imW]{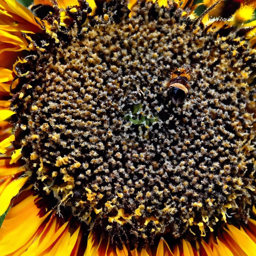}
        &\includegraphics[width=\imW]{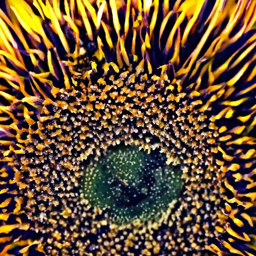}
        &\includegraphics[width=\imW]{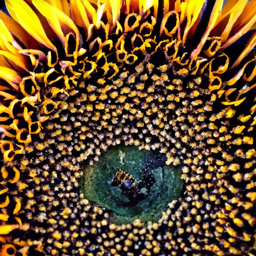}\\
         {\scriptsize (a) ChatGPT} &{\scriptsize (b) Ours} &{\scriptsize (c) Ours}
        & {\scriptsize (d) ChatGPT} &{\scriptsize (e) Ours} &{\scriptsize (f) Ours}\\
    \end{tabular}
    \caption{Images generated with our method using: \textbf{(a,d)} prompts initially generated from ChatGPT, \textbf{(b,e)} prompts improved with manual refinement, \textbf{(c,f)} same as (b,e), with one edited prompt.} 
    \label{fig:comparison_prompts}
\end{figure}

In Fig.~\ref{fig:comparison_prompts}, we compare sequences generated using the ChatGPT-generated prompts and our refined prompts (Tab.~\ref{tab:chatgpt}). The differences are usually subtle, \eg, the ChatGPT prompts for {\it Sunflower} do not align with the relative object scales, so while the zoom stack images all look plausible, the object scales in the video are jarring (though adding an extra intermediate scale solves this); but sometimes they are catastrophic, \eg, in {\it Forest}, the zoomed-out prompts describe images from viewpoints that are incompatible with other levels. 

To visualize the effects of user control, we additionally provide results with edited prompts in Fig.~\ref{fig:comparison_prompts}. In {\it Forest}, we change the innermost level from ``bark with cracks, lichen and insects" to ``a woodpecker on top of the bark", resulting in a camouflaged woodpecker (see (c), bottom). In {\it Sunflowers}, we change the outermost prompt from ``sunny day" to ``sunset time"---we see this affects all other zoom levels as well (see (f)).
We find that certain edits require changing the prompt at multiple adjacent zoom levels---otherwise coarser priors may overwhelm the creation of finer-level content (\eg, in the woodpecker example).

\section{Failure cases}
\begin{figure}[ht!]
    \centering
    \def\imW{0.5\linewidth}
    \setlength{\tabcolsep}{1pt}
     \renewcommand\cellset{\renewcommand\arraystretch{0}}%
    \renewcommand{\arraystretch}{0.5}
    \begin{tabular}{cc}
     \includegraphics[width=\imW, height=\imW]{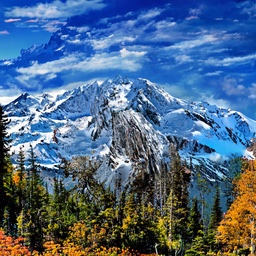}
     &\includegraphics[width=\imW, height=\imW]{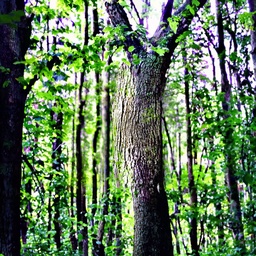}\\
     \includegraphics[width=\imW, height=\imW]{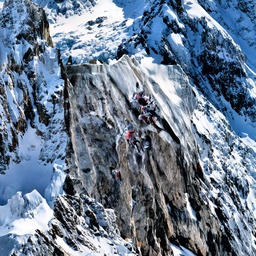}
     &\includegraphics[width=\imW, height=\imW]{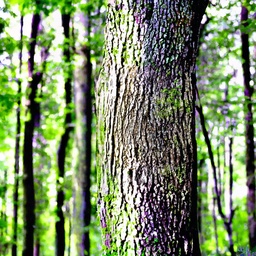}\\
     \includegraphics[width=\imW, height=\imW]{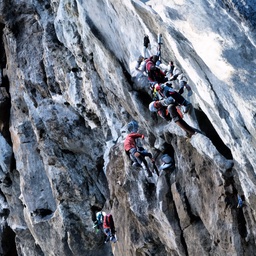}
     &\includegraphics[width=\imW, height=\imW]{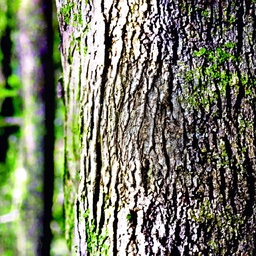}\\
    \end{tabular}
    \caption{{\bf Failure cases.} \textit{Left}: an example where the predicted images from different levels observe the scene from different viewpoints (initially from a nearly horizontal view, but finally from an oblique upward-facing view). Right: an example where image priors do not correspond to the relative scale between zoom levels, as seen in the fact that multiple scales of the bark texture exist at a single zoom level.}\label{fig:failure}
\end{figure}

Our method relies on the text-to-image diffusion model producing images of a scene at a particular set of scales from a particular viewpoint, and finding the exact set of text prompts that produce this can often be difficult. In Fig.~\ref{fig:failure}, we show examples of cases where (1) the relative scale between a set of layers does not match the distribution of images that the model intends to create, and (2) the model intends to create images from different viewpoints across different zoom levels. As mentioned in the main paper, one possible improvement could be to optimize for suitable geometric transformations between successive zoom levels. These transformations could include translation, rotation, and even scale, to find better alignment between the zoom levels and the prompts.

\clearpage
\begin{table}
    \centering
    {\small
    \begin{tabular}{c}
    \toprule
    \makecell[lt]{\tabitem {\it A straight road in the middle with alpine forests on the}\\ {\it~~~~sides under the blue sky with clouds; autumn season}}\\
    \makecell[lt]{\tabitem {\it A photo capturing the tranquil serenity of a secluded}\\{\it~~~~alpine forest road with Mount Rainier in the far end; blue}\\{\it~~~~sky; autumn season}} \\
    \makecell[lt]{\tabitem {\it A photo of serene alpine meadows against the massive }\\{\it~~~~Mount Rainier}} \\
    \makecell[lt]{\tabitem {\it Extreme close-up of the steep cliffs and rocky outcrops}\\{\it~~~~of a snow mountain occupying the entire image;}\\{\it~~~~tight framing}}\\
    \makecell[lt]{\tabitem {\it Extreme close-up of the steep cliffs and rocky outcrops}\\{\it~~~~of a snow mountain occupying the entire image;}\\{\it~~~~tight framing}}\\
    \makecell[lt]{\tabitem {\it A team of climbers with red clothes climbing on the}\\{\it~~~~rugged cliffs; low camera angle}} \\
    \bottomrule
    \end{tabular}
    }
    \caption{Complete prompts for the {\bf Mount Rainier} example (column 4 in Fig. 7) with relative scale $p=2$.}\label{tab:rainier}
\end{table}

\begin{table}
    \centering
     {\small
    \begin{tabular}{c}
    \toprule
    \makecell[lt]{\tabitem {\it Small galaxy far away surrounded by large starry dark}\\{\it~~~~sky, millions of sparkling stars against dark background}\\{\it~~~~and vast emptiness}}\\
    \makecell[lt]{\tabitem \it Beautiful, high quality  photo of Andromeda Galaxy} \\
    \makecell[lt]{\tabitem \it Galactic core, tight framing} \\
    \makecell[lt]{\tabitem \it Galactic core, tight framing} \\
    \makecell[lt]{\tabitem \it Thousands of stars against dark space in the background} \\
    \makecell[lt]{\tabitem \it Dark starry sky} \\
    \makecell[lt]{\tabitem \it Dark starry sky with a foreign solar system in the middle} \\
    \makecell[lt]{\tabitem {\it Far view of alien solar system with a star and multiple}\\ {\it~~~~exoplanets. Smaller stars in the background}} \\
    \makecell[lt]{\tabitem \it Alien solar system with one of the exoplanets in the center} \\
    \makecell[lt]{\tabitem \it  An exoplanet of a foreign solar system} \\
    \makecell[lt]{\tabitem {\it A close-up of an exoplanet in a foreign solar system,}\\ {\it~~~~revealing a dry and arid climate}}\\
    \makecell[lt]{\tabitem {\it  Very high up top-down aerial image of deserted }\\ {\it~~~~continents with reddish-hued soil in an alien planet} \\ {\it~~~~revealing a dry and arid climate}}\\
    \makecell[lt]{\tabitem {\it  High up top-down aerial image of deserted }\\ {\it~~~~continents with reddish-hued soil in an alien planet} \\ {\it~~~~revealing a dry and arid climate}}\\
    \makecell[lt]{\tabitem {\it Top-down photorealistic aerial image of a continent}\\ {\it~~~~ with a lot of deserts in an alien planet}}\\
    \makecell[lt]{\tabitem {\it Top-down photorealistic aerial image of a desert}\\ {\it~~~~  with an alien outpost in the middle}}\\
    \makecell[lt]{\tabitem {\it Top-down view of an alien outpost as seen directly above}}\\
    \bottomrule
    \end{tabular}
    }
    \caption{Complete prompts for the {\bf Galaxy} example (column 1 in Fig. 7) with relative scale $p=2$.}\label{tab:galaxy}
\end{table}

\begin{table}
    \centering
    {\small
    \begin{tabular}{c}
    \toprule
    \makecell[lt]{\tabitem {\it A girl is holding a maple leaf in front of her face, partially}\\ {\it~~~~obscuring it}}\\
    \makecell[lt]{\tabitem {\it A brightly colored autumn maple leaf. The leaf is a rich}\\{\it~~~~blend of red and yellow hue and partially covering the}\\{\it~~~~face behind it; tight framing}} \\
    \makecell[lt]{\tabitem \it A brightly colored autumn maple leaf} \\
    \makecell[lt]{\tabitem {\it Orange maple leaf texture with lots of veins;}\\{\it~~~~macrophotography}}\\
    \makecell[lt]{\tabitem {\it Macrophotograph showing the magnified veins pattern on}\\{\it~~~~the orange maple leaf texture; macrophotography}} \\
    \makecell[lt]{\tabitem {\it High resolution macrophotograph showing the magnified}\\{\it~~~~veins pattern on the orange maple leaf texture;} \\{\it~~~~macrophotography}} \\
    \bottomrule
    \end{tabular}
    }
    \caption{Complete prompts for the {\bf Maple Leaf} example (column 2 in Fig. 6) with relative scale $p=2$.}\label{tab:leaf}
\end{table}

\begin{table}
    \centering
    {\small
    \begin{tabular}{c}
    \toprule
    \makecell[lt]{\tabitem {\it An aerial view of a man lying on the picnic blanket with}\\ {\it~~~~his hand in the center of the image}}\\
    \makecell[lt]{\tabitem {\it A close-up realistic photo showing the back side of a}\\{\it~~~~men's hand; uniform lighting; this lying person's hand}\\{\it~~~~should be put on top of light faded white shirt}} \\
    \makecell[lt]{\tabitem {\it A close-up photo capturing the surface of skin of the back}\\{\it~~~~hand; uniform lighting}} \\
    \makecell[lt]{\tabitem {\it Photo taken through a light microscope of skin's }\\{\it~~~~epidermal layer. The outermost layer, the stratum corneum,}\\{\it~~~~becomes apparent; Multiple rows of dense tiny skin cells}\\{\it~~~~becomes visible in the middle.} }\\
    \makecell[lt]{\tabitem {\it Photo taken through a light microscope of a close up of}\\{\it~~~~skin's epidermal layer consisting multiple rows of dense}\\{\it~~~~tiny skin cells}} \\
    \makecell[lt]{\tabitem {\it Photo taken through a light microscope showcasing}\\{\it~~~~several skin cells with similar sizes;} \\{\it~~~~with one cell in the center
    }} \\
    \makecell[lt]{\tabitem {\it Photo taken through a light microscope  of a single round}\\{\it~~~~skin cell with its nucleus in the center}} \\
    \makecell[lt]{\tabitem {\it Photo taken through a light microscope  of a nucleus}\\{\it~~~~within a single cell}} \\
    \bottomrule
    \end{tabular}
    }
    \caption{Complete prompts for the {\bf Hand} example (column 1 in Fig. 6) with relative scale $p=4$.}\label{tab:hand}
\end{table}

\begin{table*}
    \centering
     {\small
    \begin{tabular}{c c}
     \toprule
     {\small ChatGPT generated}   &{\small Manually refined} \\
    \toprule
    \multicolumn{2}{c}{{\bf Forest}, $p=2$}\\
      \makecell[lt]{\tabitem \it View of a vast forest from a hilltop}
     & $<$level removed in refinement$>$
    \\
    \makecell[lt]{ \tabitem \it Path leading to the dense forest from open land}
    &\makecell[lt]{\tabitem \it Path leading to the dense forest from open land}
    \\  
    \makecell[lt]{\tabitem {\it Entrance of a forest} \\ {\it ~~~~with sunlight filtering through the trees}}
    &\makecell[lt]{\tabitem {\it  Entrance of a forest leading into an oak tree in the middle} \\ {\it ~~~~with sunlight filtering through the trees}}
   \\ 
     \makecell[lt]{\tabitem \it  Heart of a forest \\ \it ~~~~filled with tree trunks, leaves, vines, and undergrowth} 
    &\makecell[lt]{\tabitem  {\it Heart of a forest with a tall oak tree in the middle, }\\ {\it ~~~~filled with tree trunks, leaves, vines, and undergrowth}}
   \\
     \makecell[lt]{\tabitem  \it Single oak tree towering above the rest of the forest}
    &\makecell[lt]{ {\tabitem  \it Textured tree trunk of a tall oak tree in the middle of a forest}}
    \\
   \makecell[lt]{\tabitem \it Close-up of a textured oak tree trunk and branches}
   &\makecell[lt]{\tabitem \it Close-up of a textured oak tree trunk in a forest}
   \\
    $<$level added in refinement$>$
    &\makecell[lt]{\tabitem  \it Close-up of a textured oak tree trunk in a forest}
   \\
   \makecell[lt]{\tabitem \it Detailed view of an oak tree bark showing ridges and groove}
   &\makecell[lt]{\tabitem \it Detailed view of an oak tree bark showing ridges and groove}
   \\
   \makecell[lt]{\tabitem  \it Close-up of tree bark showing small cracks, lichen, and insects}
   &\makecell[lt]{\tabitem \it Close-up of tree bark showing small cracks, lichen, and insects}\\
   \hline 
   \multicolumn{2}{c}{{\bf Hawaii}, $p=2$}\\
   \makecell[lt]{\tabitem {\it An aerial photo capturing Hawaii's islands surrounded } \\{\it ~~~~by the vast Pacific Ocean from above} }
   &\makecell[lt]{\tabitem {\it A aerial photo capturing Hawaii's islands surrounded } \\{\it ~~~~by the vast Pacific Ocean from above } }\\
   \makecell[lt]{\tabitem {\it An aerial photo showcasing Hawaii's rugged coastlines} \\{\it ~~~~and pristine beaches } }
   &\makecell[lt]{\tabitem {\it An aerial photo showcasing Hawaii's rugged coastlines} \\{\it ~~~~and pristine beaches } }\\
   \makecell[lt]{\tabitem {\it An aerial photo revealing Hawaii's majestic mountains } \\{\it ~~~~ and lush rainforests} }
   &\makecell[lt]{\tabitem {\it An aerial photo revealing Hawaii's majestic mountains } \\{\it ~~~~ and lush rainforests} }\\
    \makecell[lt]{\tabitem {\it An aerial shot of Hawaii's dramatic crater ridges} \\{\it ~~~~ and expansive lava fields} }
   & \makecell[lt]{\tabitem {\it An aerial shot of Hawaii's dramatic crater ridges} \\{\it ~~~~ and expansive lava fields} }\\
   \makecell[lt]{\tabitem {\it Aerial view of surreal steam vents and sulphuric fumaroles} \\{\it ~~~~ within Hawaii's volcanic landscape} }
   & \makecell[lt]{\tabitem {\it An aerial close-up photo of the volcano's caldera} }\\
   \makecell[lt]{\tabitem {\it Aerial perspective capturing the raw power and } \\{\it ~~~~ natural beauty of the volcano's caldera} }
   & \makecell[lt]{\tabitem {\it An aerial close-up photo of the rim of a volcano's caldera, } \\{\it ~~~~ with a man standing on the edge.} }\\
   $<$level added in refinement$>$
   &\makecell[lt]{\tabitem {\it A top down shot of a man standing on the edge of } \\{\it ~~~~a volcano's caldera, waving at the camera.} }\\
   \hline
   \multicolumn{2}{c}{{\bf Sunflowers}, $p=2$}\\
   \makecell[lt]{\tabitem {\it A sunflower field from afar}}
   &\makecell[lt]{\tabitem {\it A sunflower field from afar}}\\
   $<$level added in refinement$>$ 
   &\makecell[lt]{\tabitem {\it A sunflower field}}\\
   \makecell[lt]{\tabitem {\it Move closer to the sunflower field; individual sunflowers} \\{\it ~~~~ becoming more defined, swaying gently in the breeze}}
   &\makecell[lt]{\tabitem {\it Close-up of rows of sunflowers of the same size facing front} \\{\it ~~~~and swaying gently in the breeze; with one in the center} }\\ 
   \makecell[lt]{\tabitem {\it  Zooms in on a specific sunflower at the field's edge} }
   &\makecell[lt]{\tabitem {\it Zooms in on a single front-facing sunflower  } \\{\it~~~~in the center at the field's edge}}\\
    \makecell[lt]{\tabitem {\it Closer view of the sunflower. Emphasize } \\{\it ~~~~the sunflower's golden petals and the intricate details} }
   & \makecell[lt]{\tabitem {\it Closer view of the sunflower in the center. Emphasize} \\{\it ~~~~the sunflower's golden petals and the intricate details} }\\
   \makecell[lt]{\tabitem {\it An image focusing solely on the center of the sunflower} \\{\it ~~~~Showcase the dark, velvety disc florets,} \\{\it ~~~~and capture the honey bee sipping nectar }\\{\it~~~~and transferring pollen} }
   &  \makecell[lt]{\tabitem {\it An extreme close-up of the center of the sunflower} \\{\it~~~~Showcase the dark, velvety disc florets,} \\{\it~~~~and capture the honey bee sipping nectar}\\{\it~~~~and transferring pollen} }\\
   \hline
    \multicolumn{2}{c}{{\bf Earth}, $p=4$}\\
   \makecell[lt]{\tabitem {\it A distant view of Earth, showing continents and oceans}}
   &\makecell[lt]{\tabitem {\it Satellite image of the Earth's surface showing} \\ {\it ~~~~a landmass in the middle as seen from space
   }}\\
   \makecell[lt]{\tabitem {\it  Zooming in on a continent, with major geographical features visible}}
   &\makecell[lt]{\tabitem {\it Satellite image of landmass of the Earth's foggy surface}} \\
   \makecell[lt]{\tabitem {\it A focused view on a specific region,} \\{\it~~~~highlighting rivers and landscapes}}
   &\makecell[lt]{\tabitem {\it Satellite image of a state in the U.S., showing the state's} \\{\it ~~~~ natural beauty with rivers, forests, and towns scattered across} }\\ 
   \makecell[lt]{\tabitem {\it Narrowing down to a dense forest area,}\\{\it~~~~showcasing the canopy and terrain} }
   &\makecell[lt]{\tabitem {\it Satellite image of a quaint American countryside surrounded} \\ {\it~~~~by forests and rivers in a foggy morning}}\\
    \makecell[lt]{\tabitem {\it Zooming in on a specific lake, surrounded by the forest.}}
   & \makecell[lt]{\tabitem {\it Satellite image  of a foggy forest with a lake in the middle } \\{\it ~~~shoot directly from above } }\\
   \makecell[lt]{\tabitem {\it Close-up of the lake's surface, with surrounding vegetation}}
   &  \makecell[lt]{\tabitem {\it Satellite image  of a lake surrounded by a forest}\\{\it~~~~shoot directly from above}}\\
   \makecell[lt]{\tabitem {\it Top-down view of a person kayaking in the lake, amidst the forest.
    }}
   &\makecell[lt]{\tabitem {\it Top down view  of a lake with a person kayaking} \\{\it~~~~shoot directly from above}}\\
    \bottomrule
    \end{tabular}
    }
    \caption{Generated prompts from ChatGPT vs. our manually refined prompts. We (1) removed prompts which are view inconsistent with others, (2) add more levels to make the relative scale correct, (3) add description to give more context about the entire scene.}
    \label{tab:chatgpt}
    \vspace{-10pt}
\end{table*}
\end{document}